\documentclass[11pt]{article}

\usepackage[final]{acl}

\usepackage{times}
\usepackage{latexsym}

\usepackage[T1]{fontenc}

\usepackage[utf8]{inputenc}

\usepackage{microtype}

\usepackage{inconsolata}

\usepackage{graphicx}
\usepackage{amsmath}
\usepackage{amssymb}
\usepackage{array}
\usepackage{arydshln}
\usepackage{booktabs}
\usepackage{enumitem}
\usepackage{fvextra}
\usepackage{hyperref}
\usepackage{multicol}
\usepackage{multirow}
\usepackage[normalem]{ulem}
\usepackage{pifont}
\usepackage{ragged2e}
\usepackage{soul}
\usepackage{subcaption}
\usepackage{tabularx}
\usepackage{tikz}
\usepackage{xurl}
\usepackage{xcolor}

\definecolor{lightgrey}{rgb}{0.9,0.9,0.9}
\definecolor{lightred}{RGB}{255,220,220}
\definecolor{lightblue}{RGB}{220,235,255}

\newcommand{\setreturn}[1]{
  \hypertarget{return:#1}{}
}

\DeclareRobustCommand{\backtomain}[1]{%
  \hyperlink{return:#1}{\uline{Return to main text.}}%
}

\DeclareRobustCommand{\backtoappendix}[1]{%
  \hyperlink{return:#1}{\uline{Return to Appendix.}}%
}

\usetikzlibrary{calc}

\newcommand{\DeltaSetRange}[2]{
  \def\DeltaMin{#1}
  \def\DeltaMax{#2}
}

\newcommand{\DeltaSetRangeBLEU}[2]{
  \def\DeltaMinBLEU{#1}
  \def\DeltaMaxBLEU{#2}
}

\newcommand{\DeltaSetRangeROUGE}[2]{
  \def\DeltaMinROUGE{#1}
  \def\DeltaMaxROUGE{#2}
}

\newcommand{\DeltaSetRangeBERT}[2]{
  \def\DeltaMinBERT{#1}
  \def\DeltaMaxBERT{#2}
}

\newlength{\DeltaColW}
\newcommand{\DeltaHeat}[1]{
  \begingroup
  \pgfmathsetmacro{\pct}{
    min(100,max(0,100*(#1 - \DeltaMin)/(\DeltaMax - \DeltaMin)))
  }
  \tikz[baseline=(t.base)]{
    \node[inner xsep=2pt, inner ysep=1pt, minimum width=\DeltaColW, anchor=base, fill=teal!\pct!white] (t) {\makebox[\DeltaColW][c]{#1}};
  }
  \endgroup
}

\newcommand{\DeltaHeatBLEU}[1]{
  \begingroup
  \pgfmathsetmacro{\pct}{
    min(100,max(0,100*(#1 - \DeltaMinBLEU)/(\DeltaMaxBLEU - \DeltaMinBLEU)))
  }
  \tikz[baseline=(t.base)]{
    \node[inner xsep=2pt, inner ysep=1pt, minimum width=\DeltaColW, anchor=base, fill=teal!\pct!white] (t) {\makebox[\DeltaColW][c]{#1}};
  }
  \endgroup
}

\newcommand{\DeltaHeatROUGE}[1]{
  \begingroup
  \pgfmathsetmacro{\pct}{
    min(100,max(0,100*(#1 - \DeltaMinROUGE)/(\DeltaMaxROUGE - \DeltaMinROUGE)))
  }
  \tikz[baseline=(t.base)]{
    \node[inner xsep=2pt, inner ysep=1pt, minimum width=\DeltaColW, anchor=base, fill=teal!\pct!white] (t) {\makebox[\DeltaColW][c]{#1}};
  }
  \endgroup
}

\newcommand{\DeltaHeatBERT}[1]{
  \begingroup
  \pgfmathsetmacro{\pct}{
    min(100,max(0,100*(#1 - \DeltaMinBERT)/(\DeltaMaxBERT - \DeltaMinBERT)))
  }
  \tikz[baseline=(t.base)]{
    \node[inner xsep=2pt, inner ysep=1pt, minimum width=\DeltaColW, anchor=base, fill=teal!\pct!white] (t) {\makebox[\DeltaColW][c]{#1}};
  }
  \endgroup
}

\title{Judgment-Grounded Expansion for Peer Review Generation}

\author{
LU Sheng\textsuperscript{1}, Lizhen Qu\textsuperscript{2}, Iryna Gurevych\textsuperscript{1} \\ [0.3cm]
\textsuperscript{1}Ubiquitous Knowledge Processing Lab (UKP Lab), Department of Computer Science, \\
Technical University of Darmstadt and National Research Center for Applied Cybersecurity ATHENE \\
\textsuperscript{2}Department of Data Science and Artificial Intelligence (DSAI), Monash University \\
\texttt{\small www.ukp.tu-darmstadt.de}}

\begin{document}
\maketitle
\begin{abstract}

Automatic review generation is a promising direction for accelerating scientific progress. While most work adopts an end-to-end setup, its fully automated nature makes it less suitable for settings that demand accountability. To better balance automation and accountability, we formalize \emph{judgment-grounded expansion}, a human-AI collaboration mode where a reviewer provides an evaluative claim and the system expands it into review comment candidate(s). We model it as a structured generate-check-refine process and conduct a user study to collect human-model interaction data. We study two practical challenges for judgment-grounded expansion: scalable evaluation and candidate set curation. We develop methods to simulate the process for large-scale evaluation, and show that conformal prediction is well suited to balancing candidate set size and target coverage. Our work establishes judgment-grounded expansion as a concrete task and provides empirical and methodological foundations for the design of future collaborative review generation systems.\footnote{Our code and data are available \href{https://anonymous.4open.science/r/judgment-grounded-expansion-B28D/}{here}.}

\end{abstract}

\section{Introduction}

Scientific progress relies on timely peer review. However, the academic publishing system is under increasing pressure from the rapid growth of submissions, creating a bottleneck that may slow scientific iteration \cite{schulz_is_2022,yuan_can_2022,biswas_chatgpt_2023,lin_automated_2023}. There has been rising interest in automating the peer review process. In particular, automatic review generation has emerged as a promising direction with the potential to support faster scientific iteration \cite{kuznetsov_what_2024,liang_can_2024,chen_ai4research_2025,zhu_deepreview_2025,zhuang_large_2025}.

\begin{figure}[!t]
\centering
\includegraphics[width=0.98\linewidth]{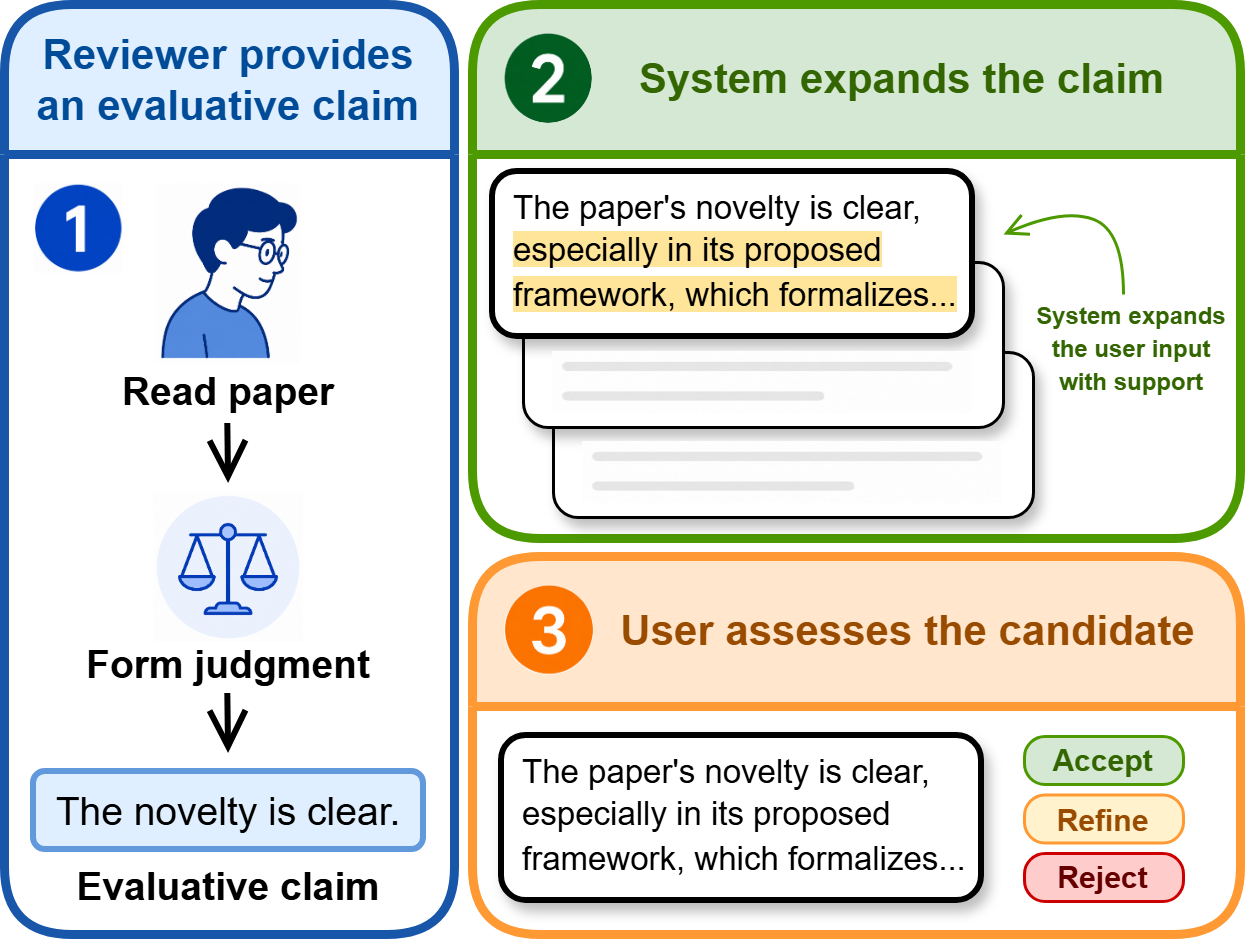}
\caption{Judgment-grounded expansion encourages the reviewer's engagement and supports accountability.}
\label{teaser}
\end{figure}

Most existing automatic review generation systems adopt an end-to-end setup, autonomously generating a review with little or no human intervention during generation \cite{gao_reviewagents_2025,idahl_openreviewer_2025,zhu_deepreview_2025}. While full automation minimizes human effort, it limits human control during content creation and raises accountability concerns: studies show that fully automated reviews become homogenized across different papers \cite{lu_identifying_2025,baumann2026stop}, exhibit verification failures \cite{dycke2026counterfactual,you2026preventing}, and can be susceptible to manipulation \cite{ye_are_2024}.

Human-AI collaboration provides a compelling alternative \cite{yang_human-ai_2024,dutta2025haico,huang_how_2025}. Writing expansion is a major interaction mode in collaborative review generation: the reviewer provides a draft fragment, and the system expands it into a more complete review comment \cite{chen_envisioning_2025}. It is a promising direction for settings that demand accountability, as it maintains a degree of human control while delegating part of the work to the system. However, the interaction pipeline of writing expansion remains under-specified, especially in terms of human input and the quality check mechanism, which is crucial to the degree of human control during generation.

We move beyond general writing expansion and formalize a more specific collaboration mode: \emph{judgment-grounded expansion}. In this setup, the human reviewer provides an evaluative claim regarding the paper and the system expands it with support grounded in the submission, producing one or several review comment candidates for the reviewer to inspect (see Figure \ref{teaser}). It encourages the reviewer's active engagement: the reviewer initiates the process with an evaluative claim, remains responsible for the evaluative stance, and controls output quality through \emph{accepting}, \emph{rejecting}, or \emph{refining} the candidate over multiple rounds. Anchoring generation in the reviewer's own evaluative claim is a key specification of judgment-grounded expansion, which better supports accountability in formal settings such as conference reviewing.


To study judgment-grounded expansion in practice, we conduct a user study and collect human-model interaction data. Our study characterizes reviewer input, quality check behavior, and user experience, which provides a resource for research on judgment-grounded expansion.

Moreover, we study two practical challenges for judgment-grounded expansion. The first is \emph{scalable evaluation}: reproducible large-scale evaluation is essential for assessing design choices such as backbone model selection, yet judgment-grounded expansion requires human participation which is difficult to scale. We address this by developing methods to simulate judgment-grounded expansion. The second is \emph{candidate set curation}: when the system presents multiple candidates to improve coverage of the reviewer's intended comment, it should keep the candidate set compact to avoid excessive reviewer burden. This size-coverage trade-off aligns naturally with conformal prediction \cite{angelopoulos2023gentle}, which constructs prediction sets carrying a user-specified coverage guarantee, keeping sets compact when the model is confident and enlarging them only when the model is uncertain. We accordingly cast candidate set curation as a conformal prediction problem.

Our contributions are summarized as follows. \textbf{(1) New task formulation.} We formalize and model judgment-grounded expansion as a generate-check-refine process. \textbf{(2) New interaction data.} We conduct a user study and collect interaction data, providing a resource for studying judgment-grounded expansion. \textbf{(3) New evaluation pipeline.} We develop a novel simulation method that enables reproducible evaluation at scale.

\section{Related work}

\subsection{Review generation}

End-to-end review generation is an active line of research in peer review automation. Existing approaches largely fall into three categories: (1) prompt-based \cite{du_llms_2024,liang_can_2024}; (2) agent-based \cite{chamoun_automated_2024,jin_agentreview_2024,lu_ai_2024,gao_reviewagents_2025}; (3) fine-tuned LLM \cite{faizullah_limgen_2024,tan_peer_2024,idahl_openreviewer_2025,zhu_deepreview_2025}. With system prompts or pre-configured modules, they autonomously generate reviews with little or no human intervention. While it largely removes the need for human involvement, it places control over the review in the model during generation, which limits accountability. This necessitates human-AI collaborative generation, where reviewers remain involved in the generation process and can steer the system toward more accountable content.


Human-AI collaboration has gained growing attention in review generation \cite{sun_reviewflow_2024,sun_metawriter_2024}. Writing expansion, a major interaction mode in collaborative review generation \cite{chen_envisioning_2025,sadeghian2026peerprism}, is particularly suitable for settings that demand accountability, as it keeps reviewers involved while delegating part of the work to the system. However, the key components of writing expansion, such as human input and quality check, remain under-specified. This gap motivates judgment-grounded expansion, a more specific collaboration mode aimed at better balancing automation and accountability.

\subsection{Review evaluation}
\label{what_is_a_good_review}

Evaluating human-AI collaboration ideally involves human participation. While being essential for capturing subjective experience \cite{lee_evaluating_2024,fragiadakis_evaluating_2025}, it limits reproducibility \cite{belz_missing_2023,belz_non-repeatable_2023} and scalability, especially for review writing, a costly task where participants must be domain experts and engage in the full reviewing process. Reproducible large-scale evaluation is essential for understanding how different task and model factors affect generation. We enable this by developing methods to simulate judgment-grounded expansion.

\setreturn{review_evaluation}
For evaluating generated review content, prior studies consider several desiderata, and some are operationalized with metrics \cite{ramachandran_automated_2017,yuan_can_2022,sizo_defining_2025}. Many studies measure alignment with human-written reviews using text similarity metrics. Beyond surface similarity, human evaluation and LLM-as-a-judge \cite{chiang_can_2023,zheng_judging_2023} provide more nuanced assessments for qualities that are hard to quantify, such as constructiveness. Recent work further emphasizes two additional concerns: hallucination \cite{bang_multitask_2023,mckenna_sources_2023,zhang_how_2024} and adversarial robustness \cite{ye_are_2024,zhu_deepreview_2025}. In our evaluation, we adopt these commonly used methods (summarized in Table \ref{evaluation_methods_related_work}) and assess hallucination and adversarial robustness.

\subsection{Conformal prediction}

Conformal prediction provides a principled way to construct variable-size prediction sets with a target coverage guarantee \cite{angelopoulos2023gentle,angelopoulos2024conformal}. It has been increasingly applied to natural language processing tasks \cite{campos2024conformal,huang2024conformalrank,quach2024conformal,sheng2025analyzing,vishwakarma2025prune}. Conformal prediction is a natural fit for judgment-grounded expansion: set prediction is a useful strategy to increase the likelihood of covering the reviewer-intended review comment; however, the candidate set must remain compact to avoid excessive reviewer burden. We investigate conformal prediction as a promising solution to this size and coverage trade-off.

\section{Judgment-grounded expansion}
\label{key_point_and_judgment_expansion}

\begin{figure}[!ht]
\centering
\vspace{0.2em}
\includegraphics[width=0.828\linewidth]{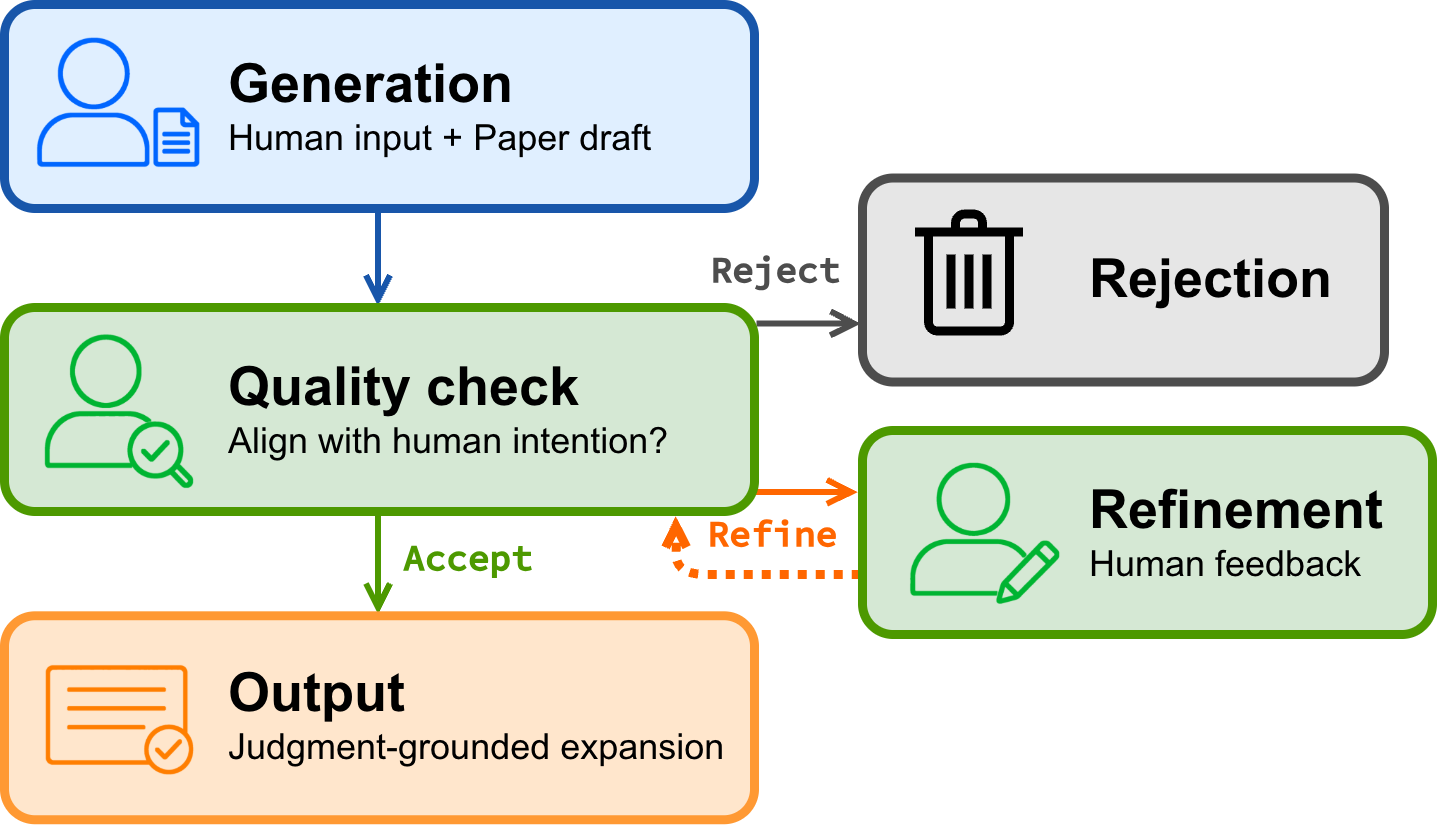}
\caption{Our proposed framework.}
\label{overall_framework}
\vspace{-0.86\baselineskip}
\end{figure}

We model judgment-grounded expansion as a generate-check-refine process (see Figure \ref{overall_framework}).

\subsection{Human input}
\label{human_input}

The input should steer the generation by the reviewer's own evaluative claim. We therefore define the human input as, at minimum, a key point (e.g., \emph{Novelty}) and its corresponding judgment (e.g., \emph{strength}). This aligns with \citet{hua_argument_2019}'s view of review as arguments each consisting of a key point, a judgment, and support. 

\subsection{Generation}

The review generation system takes the human input and the paper as inputs, and outputs either a single review comment or a set of candidates. Compared to single-point prediction, set prediction is potentially more practical, as a single generated comment may not always match the reviewer's intended comment. Set prediction better supports judgment-grounded expansion by exploiting the generative capacity of models to explore multiple plausible expansions, increasing the likelihood that the candidate set covers the user's intention.

\subsection{Quality check}

After receiving the generated comment(s), the human reviewer performs quality check by assessing whether it aligns with their intention. If it passes the check, the workflow proceeds directly to \emph{Output}. Otherwise, the human reviewer initiates \emph{Refinement} or stops the interaction by rejecting the generation.


We define two high-level categories of quality check dimensions, \textbf{Intention} and \textbf{Grounding}, which reflect the key requirements of writing expansion: the system must interpret the reviewer's intent, and provide support to the reviewer's input. The quality check is carried out progressively, which isolates issues early and allows the reviewer to intervene in a systematic manner.

Passing the quality checks does not mean the generation is immediately ready for use. It marks the completion of judgment-grounded expansion. Additional revisions, such as tone adjustment, are handled outside the scope of this framework.


\subsubsection{Intention}
\label{intention}

This dimension assesses whether the system correctly interprets the human input and captures their intention. It involves \textbf{Relevance}, whether the generation addresses the intended key point and judgment, and \textbf{Specification}, whether the generation captures the intended specificity of the input.

\subsubsection{Grounding}
\label{grounding}

This dimension concerns whether the generated review provides sufficient support for the human input. It includes \textbf{Evidence}, which evaluates whether the generated evidence supports the human input. The \emph{Evidence} check inspects: \uline{hallucination}, whether the generated evidence is hallucinated; \uline{match}, whether the generated evidence matches the evidence the human reviewer would use (a mismatch does not necessarily mean an error); \uline{sufficiency}, when the generated evidence differs from what a human reviewer would use, whether it is still sufficient to support the human input. \textbf{Reasoning} assesses whether the reasoning is sufficiently developed to support the human input. A generation may fail this criterion if it lacks a clear logical connection to the human input.

\subsection{Refinement}

\setreturn{user_study}
If the human reviewer chooses to refine the generation, they may take different actions depending on the specific issue identified. If the generation fails \textbf{Relevance}, the human reviewer may reiterate the original input to re-generate. If it fails \textbf{Specification}, the reviewer can clarify by providing a more detailed input. For \textbf{Evidence}, if \uline{hallucination} is detected, the reviewer can request the system to generate alternative evidence; if \uline{sufficiency} is the issue, the reviewer may ask the system to generate stronger or additional evidence. If the generation fails \textbf{Reasoning}, the human reviewer can prompt the system to elaborate the argumentation.

\section{User study}
\label{user_study}

Based on our proposed framework, we prototype an interface to support judgment-grounded expansion and conduct a user study. We recruited 6 participants, and collected 81 human-model interaction instances for 14 papers (see Appendix \ref{more_on_user_study} for more details). We observe that:

\vspace{0.2em}

\noindent\textbf{(1) Concise input supports successful expansion.} User inputs average 17.5 words, with 88.9\% containing fewer than 30 words. 45.7\% contain only a key point and evaluative stance (e.g., ``The novelty of this paper is good.''), 34.5\% additionally reference specific paper elements (e.g., a section or table), and 19.8\% include suggestions or recommendations. 72.8\% of interactions result in immediate acceptance, suggesting that the system can often operate with concise user input.

\vspace{0.2em}

\noindent\textbf{(2) Failure cases reveal input requirements.} 16.1\% of interactions result in rejection. We observe 3 recurring user input patterns in these cases: (1) 33.3\% are too brief (less than 10 words) or generic; (2) 25.0\% are hedged inputs that express uncertainty or pose questions; (3) 16.7\% reference external papers, which our interface cannot access.\footnote{We intentionally disable web access to prevent the system from retrieving publicly available reviews.} 11.1\% of interactions involve refinement before acceptance. In these cases, users mainly requested additional content to be incorporated, or clarified the intended focus. These patterns suggest that effective judgment-grounded expansion requires user input to be clear, definitive, and specific.

\vspace{0.2em}

\noindent\textbf{(3) User feedback highlights value and trade-offs.} We collect participants’ subjective experience through a post-study questionnaire. On 7-point Likert-scale questions, participants reported moderate effort in correcting generated reviews before submission (Mean=4.15) and satisfaction with the final reviews (Mean=5.69). Participants also indicated a preference for judgment-grounded expansion over writing reviews from scratch (Mean=5.92). Participants reported in open-ended responses that judgment-grounded expansion was more efficient than writing from scratch. One participant noted that it helped validate straightforward judgments quickly and reallocate effort to more uncertain cases. They felt responsible for the final review because the claim came from their own judgment. Participants also found multiple candidates helpful, but they introduced some additional cognitive burden.

\section{Simulation and evaluation}

We simulate the judgment-grounded expansion process to enable large-scale, reproducible evaluation, which is valuable for understanding the effect of task and model factors, but impractical to conduct with human reviewers.

\subsection{Simulation}
\label{simulating_collaborative_generation}

\vspace{0.3em}

\setreturn{simulating_human_input}
\phantomsection\label{simulating_human_input}
\noindent\textbf{Human input.}\ \ We design two methods to approximate human input: \textbf{(1) Exemplar-based proxy.}\ \ We use real human inputs collected from our user study as in-context exemplars to generate simulated human inputs (see Table \ref{creating_exemplar_based_proxy} for the prompt). This produces inputs that reflect the characteristics of real reviewer inputs as summarized in \S\ref{user_study}. This proxy is suitable for simulation that mimics specific human input styles, but requires a sufficient pool of exemplars. \textbf{(2) Aspect-based proxy.}\ \ We use aspect and the \emph{strength} and \emph{weakness} labels in the review as a proxy of reviewer judgment. Aspects are characteristics of a paper that a reviewer makes judgment on. We identify aspects of human-written reviews using the tagger from \citet{lu_identifying_2025}, which provides both \textsc{coarse} and \textsc{fine} aspects in terms of granularity (see Table \ref{examples_of_aspects}). We begin with \textsc{coarse} aspects, and use the corresponding \textsc{fine} aspects when a generation fails the \emph{Specification} check. This proxy controls for variables such as stylistic variation. It limits leakage of reference wording while retaining the intended topic and stance signal. It is a low-information proxy as aspects are less expressive than real human input.

\vspace{0.3em}

\setreturn{quality_check_simulation}
\noindent\textbf{Quality check.}\ \ For the \emph{Relevance} and \emph{Specification} checks, we use the aspect tagger from \citet{lu_identifying_2025} to label the aspects in the generated reviews. These checks serve as an intent-alignment proxy: a generated review passes these checks only if it covers all aspects identified for the corresponding human-written review. For \emph{Evidence} and \emph{Reasoning}, we use \href{https://openai.com/index/openai-o3-mini/}{OpenAI o3-mini} to simulate the checks, and the prompt is shown in Table \ref{quality_check_prompts}.

\setreturn{review_generation}
\subsection{Evaluation}

We first perform a sanity check of our simulation (\S\ref{simulation_validation}). Then, building on our framework and simulation methods, we conduct reproducible large-scale evaluation that compares judgment-grounded expansion with an end-to-end baseline to quantify the effect of reviewer judgment (\S\ref{value_of_judgment_grounded_expansion}). We then introduce \emph{judgment-utilization score} as a diagnostic measure of how much a system benefits from reviewer judgment (\S\ref{collaborative_ability}).

\subsubsection{Experimental setup}
\label{experimental_settings}

\vspace{0.2em}

\noindent\textbf{Dataset.}\ \ We use reviews from \href{https://tudatalib.ulb.tu-darmstadt.de/handle/tudatalib/4460}{ARR} and \href{https://docs.openreview.net}{ICLR}. Both datasets contain reviews along with the paper draft and metadata such as strength/weakness tags and review scores. We remove low-quality reviews (e.g., extremely short ones). We randomly select 100 reviews from each dataset. We segment each review by key points: when reviews contain explicit structure, such as bullet points or numbered lists, we use these cues for segmentation; otherwise, we manually segment them based on content.

\vspace{0.2em}

\noindent\textbf{Review generation.}\ \ We experiment with a range of models, including vision LLMs. We also test DeepReview, an end-to-end review generation system that is shown to outperform other existing systems \cite{zhu_deepreview_2025}. We include it to examine how such a system performs with reviewer judgment. The prompts for end-to-end generation and judgment-grounded expansion are kept identical except that the latter includes reviewer judgment (see Tables \ref{end_to_end_prompt} and \ref{collaborative_prompt}). Since DeepReview is a pre-configured system, we could not modify its generation pipeline. We therefore append the judgment-grounded expansion prompt to the paper text as input. See Appendix \ref{more_on_experiment_settings} for more details.


\setreturn{sanity_check}
\subsubsection{Sanity check of the simulation}
\label{simulation_validation}

We assess the sanity of two key components in our simulation: human input and quality check. For human input, we compare the generated exemplar-based proxies with the real user inputs we collected in terms of input patterns. The proxies largely follow the patterns reported in \S\ref{user_study}: they are concise, with an average length of 7.2 words, and often contain only a key point and evaluative stance (68.5\%). Another 18.1\% reference specific paper elements, and 13.5\% include suggestions or recommendations, compared to 34.5\% and 19.8\% in real user inputs. This suggests that the exemplar-based proxy captures broad patterns of real user input, but also shows a potential bias toward the most common input style. For the aspect-based proxy, we use the aspect tagger from \citet{lu_identifying_2025}, which reports an F1 of 0.77 on its labeling task, indicating that the resulting aspect signal is sufficient to serve as a low-information proxy for human judgment. While these checks cannot directly validate the proxies against real human input, it supports their use as reasonable approximations.

We conduct human evaluation to inspect the alignment between simulated quality checks and human assessments. Table \ref{iaa_quality_check} shows the agreement for the quality check, including both human-model and human-human agreement. The model achieves reasonably strong agreement with human evaluators on \emph{Relevance}, \emph{Evidence}, and \emph{Reasoning}, and the two human evaluators also exhibit good agreement on these dimensions. \emph{Specification} shows lower human-model and human-human agreement, indicating that this dimension may be intrinsically harder to judge consistently.

We also inspect the consistency of the quality check simulation across different LLMs. We show in Figure \ref{quality_check_consistency} the simulated checks are not idiosyncratic to a specific model. Larger models ($\geq$ 32B) generally yield consistent quality check outcomes.

\begin{table*}[!ht]
\centering
\scriptsize
\renewcommand{\arraystretch}{0.78}
\setlength{\abovecaptionskip}{2pt}
\setlength{\belowcaptionskip}{2pt}
\begin{subtable}{0.49\textwidth}
\centering
\setlength{\tabcolsep}{1.12pt}
{\settowidth{\DeltaColW}{00.00}
\DeltaSetRangeBLEU{-25}{45}
\DeltaSetRangeROUGE{-1}{16}
\DeltaSetRangeBERT{-0.5}{3.5}
\begin{tabular}{lccccccccc}
\toprule
\multirow{2.6}{*}{\textbf{model}} & \multicolumn{3}{c}{\textbf{BLEU\% $\uparrow$}} & \multicolumn{3}{c}{\textbf{ROUGE-L\% $\uparrow$}} & \multicolumn{3}{c}{\textbf{BERTScore\% $\uparrow$}} \\ \cmidrule(l{2pt}r){2-4} \cmidrule(l{2pt}r){5-7} \cmidrule(l{2pt}r){8-10}
& \textbf{e2e} & \textbf{jge} & \textbf{$\Delta$\%} & \textbf{e2e} & \textbf{jge} & \textbf{$\Delta$\%} & \textbf{e2e} & \textbf{jge} & \textbf{$\Delta$\%} \\ \midrule
G-O-20B    & 2.20 & \textbf{3.00} & \DeltaHeatBLEU{+36.36} & 15.27 & \textbf{16.71} & \DeltaHeatROUGE{+9.43}  & 63.92 & \textbf{65.47} & \DeltaHeatBERT{+2.42} \\
G-O-120B   & 2.61 & \textbf{2.87} & \DeltaHeatBLEU{+9.96}  & 15.61 & \textbf{17.08} & \DeltaHeatROUGE{+9.42}  & 63.93 & \textbf{65.68} & \DeltaHeatBERT{+2.74} \\ \hdashline\noalign{\vskip 2pt}
Q3-1.7B    & \textbf{4.70} & 3.68 & \DeltaHeatBLEU{-21.70} & \textbf{18.98} & 18.86 & \DeltaHeatROUGE{-0.63} & \textbf{67.63} & 67.39 & \DeltaHeatBERT{-0.35} \\
Q3-8B      & \textbf{5.52} & 4.75 & \DeltaHeatBLEU{-13.95} & 18.90 & \textbf{19.05} & \DeltaHeatROUGE{+0.79} & 67.76 & \textbf{67.94} & \DeltaHeatBERT{+0.27} \\
Q3-14B     & \textbf{4.69} & 4.43 & \DeltaHeatBLEU{-5.54}  & 18.45 & \textbf{18.60} & \DeltaHeatROUGE{+0.81} & 67.70 & \textbf{67.72} & \DeltaHeatBERT{+0.03} \\
Q3-32B     & 4.21 & \textbf{4.69} & \DeltaHeatBLEU{+11.40} & 17.82 & \textbf{18.87} & \DeltaHeatROUGE{+5.89} & 67.15 & \textbf{67.95} & \DeltaHeatBERT{+1.19} \\ \hdashline\noalign{\vskip 2pt}
Q3-V-8B    & 3.30 & \textbf{3.68} & \DeltaHeatBLEU{+11.52} & 17.31 & \textbf{17.78} & \DeltaHeatROUGE{+2.72} & 66.45 & \textbf{66.84} & \DeltaHeatBERT{+0.59} \\
Q3-V-32B   & \textbf{3.32} & 3.21 & \DeltaHeatBLEU{-3.31}  & 16.62 & \textbf{16.69} & \DeltaHeatROUGE{+0.42} & 66.22 & \textbf{66.66} & \DeltaHeatBERT{+0.66} \\ \hdashline\noalign{\vskip 2pt}
R1-Q-14B   & 4.60 & \textbf{4.91} & \DeltaHeatBLEU{+6.74}  & 17.33 & \textbf{18.84} & \DeltaHeatROUGE{+8.71} & 66.84 & \textbf{67.85} & \DeltaHeatBERT{+1.51} \\
R1-Q-32B   & 4.18 & \textbf{4.85} & \DeltaHeatBLEU{+16.03} & 17.31 & \textbf{18.82} & \DeltaHeatROUGE{+8.72} & 66.79 & \textbf{67.89} & \DeltaHeatBERT{+1.65} \\ \hdashline\noalign{\vskip 2pt}
R1-L-8B    & \textbf{3.16} & 3.00 & \DeltaHeatBLEU{-5.06}  & 16.56 & \textbf{17.93} & \DeltaHeatROUGE{+8.27} & 65.92 & \textbf{66.74} & \DeltaHeatBERT{+1.24} \\
R1-L-70B   & 3.94 & \textbf{5.00} & \DeltaHeatBLEU{+26.90} & 17.59 & \textbf{18.87} & \DeltaHeatROUGE{+7.28} & 67.15 & \textbf{68.12} & \DeltaHeatBERT{+1.44} \\ \hdashline\noalign{\vskip 2pt}
DeepR      & 1.86 & \textbf{1.98} & \DeltaHeatBLEU{+6.45}  & 11.09 & \textbf{11.10} & \DeltaHeatROUGE{+0.09} & 66.41 & \textbf{66.50} & \DeltaHeatBERT{+0.14} \\
\bottomrule
\end{tabular}}
\caption{Text similarity}
\label{evaluation_scores_similarity}
\end{subtable}
\hfill
\begin{subtable}{0.49\textwidth}
\centering
\setlength{\tabcolsep}{3pt}
\renewcommand{\arraystretch}{0.9148}
\begin{tabular}{lcccccccc}
\toprule
\multirow{4.6}{*}{\textbf{model}} & \multicolumn{4}{c}{\textbf{Soundness}} & \multicolumn{4}{c}{\textbf{Overall assessment}} \\ \cmidrule(lr{3pt}){2-5} \cmidrule(lr{3pt}){6-9}
& \multicolumn{2}{c}{\textbf{$\leq$0.5}\% $\uparrow$} & \multicolumn{2}{c}{\textbf{MAE\%} $\downarrow$} & \multicolumn{2}{c}{\textbf{$\leq$0.5}\% $\uparrow$} & \multicolumn{2}{c}{\textbf{MAE\%} $\downarrow$} \\ \cmidrule(lr{3pt}){2-3} \cmidrule(lr{3pt}){4-5} \cmidrule(lr{3pt}){6-7} \cmidrule(lr{3pt}){8-9}
& \textbf{e2e} & \textbf{jge} & \textbf{e2e} & \textbf{jge} & \textbf{e2e} & \textbf{jge} & \textbf{e2e} & \textbf{jge} \\ \midrule
G-O-20B & \textbf{76.47} & 74.47 & 53.53 & \textbf{48.94} & 64.71 & \textbf{69.15} & 63.53 & \textbf{56.91} \\
G-O-120B & 70.84 & \textbf{76.77} & 56.77 & \textbf{51.52} & 66.67 & \textbf{76.76} & 60.42 & \textbf{47.98} \\ \hdashline\noalign{\vskip 2pt}
Q3-1.7B & \textbf{75.26} & 64.40 & \textbf{49.48} & 63.56 & \textbf{64.95} & 64.40 & 59.28 & \textbf{57.63} \\
Q3-8B & \textbf{59.79} & 53.57 & \textbf{67.01} & 69.64 & 54.64 & \textbf{55.95} & \textbf{65.46} & 68.45 \\
Q3-14B & \textbf{82.47} & 76.19 & \textbf{47.42} & 51.19 & \textbf{81.44} & 78.57 & \textbf{47.94} & 57.19 \\
Q3-32B & 61.86 & \textbf{65.22} & 61.86 & \textbf{59.24} & 48.45 & \textbf{51.09} & 74.23 & \textbf{69.02} \\ \hdashline\noalign{\vskip 2pt}
Q3-V-8B & 72.00 & \textbf{73.33 }& 55.00 & \textbf{54.00} & 57.00 & \textbf{70.66} & 63.50 & \textbf{58.00} \\
Q3-V-32B & 65.00 & \textbf{70.00 }& 61.00 & \textbf{54.00} & 52.00 & \textbf{58.00} & 70.00 & \textbf{64.50} \\ \hdashline\noalign{\vskip 2pt}
R1-Q-14B & 68.05 & \textbf{72.22} & 56.19 & \textbf{55.56} & 50.52 & \textbf{61.11} & 76.80 & \textbf{74.31} \\
R1-Q-32B & 63.44 & \textbf{65.17} & \textbf{60.22} & 60.67 & 67.74 & \textbf{70.79} & 57.53 & \textbf{55.06} \\ \hdashline\noalign{\vskip 2pt}
R1-L-8B & \textbf{55.55} & 41.47 & \textbf{76.77} & 95.12 & 39.39 & \textbf{41.47} & \textbf{95.96} & 97.56 \\
R1-L-70B & \textbf{69.39} & 69.32 & 56.12 & \textbf{51.14} & 51.02 & \textbf{59.09} & 69.90 & \textbf{64.20} \\
\bottomrule
\end{tabular}
\caption{Recommendation accuracy}
\label{evaluation_scores_score_prediction}
\end{subtable}

\vspace{0.35em}

\begin{subtable}{0.49\textwidth}
\centering
\setlength{\tabcolsep}{6.2pt}
\begin{tabular}{lcccccc}
\toprule
\multirow{2.6}{*}{\textbf{model}} & \multicolumn{2}{c}{\textbf{TA} $\uparrow$} & \multicolumn{2}{c}{\textbf{CV} $\uparrow$} & \multicolumn{2}{c}{\textbf{AD} $\uparrow$} \\ \cmidrule(lr{3pt}){2-3}\cmidrule(lr{3pt}){4-5}\cmidrule(lr{3pt}){6-7}
& \textbf{e2e} & \textbf{jge} & \textbf{e2e} & \textbf{jge} & \textbf{e2e} & \textbf{jge} \\ \midrule
G-O-20B & \textbf{64.82} & 25.93 & \textbf{72.23} & 20.99 & \textbf{68.52} & 24.69 \\
G-O-120B & \textbf{53.08} & 36.30 & \textbf{62.78} & 30.32 & \textbf{54.26} & 37.76 \\ \hdashline\noalign{\vskip 2pt}
Q3-1.7B & 47.46 & \textbf{48.31} & \textbf{55.94} & 41.53 & 44.92 & \textbf{54.24} \\
Q3-8B & \textbf{46.43} & 45.84 & \textbf{54.17} & 41.08 & 42.86 & \textbf{52.98} \\
Q3-14B & 44.26 & \textbf{46.15} & \textbf{58.10} & 36.50 & 44.94 & \textbf{47.86} \\
Q3-32B & \textbf{49.46} & 42.39 & \textbf{63.04} & 34.24 & \textbf{51.09} & 45.11 \\ \hdashline\noalign{\vskip 2pt}
Q3-V-8B & 37.84 & \textbf{50.00} & \textbf{59.46} & 32.44 & \textbf{48.65} & 45.95 \\
Q3-V-32B & 42.28 & \textbf{51.66} & \textbf{52.46} & 44.52 & 43.91 & \textbf{52.56} \\ \hdashline\noalign{\vskip 2pt}
R1-Q-14B & 29.17 & \textbf{66.67} & 31.95 & \textbf{65.28} & 25.70 & \textbf{71.53} \\
R1-Q-32B & 19.10 & \textbf{77.53} & 26.97 & \textbf{71.91} & 19.67 & \textbf{76.97} \\ \hdashline\noalign{\vskip 2pt}
R1-L-8B & \textbf{54.46} & 44.30 & \textbf{63.21} & 36.80 & \textbf{53.08} & 46.93 \\
R1-L-70B & 19.54 & \textbf{75.86} & 22.99 & \textbf{74.71} & 16.09 & \textbf{82.19} \\ \hdashline\noalign{\vskip 2pt}
DeepR & 42.90 & \textbf{47.29} & 45.66 & \textbf{45.93} & \textbf{47.06} & 45.23 \\ \midrule
\textsc{average} & 41.70 & \textbf{51.11} & \textbf{50.07} & 45.65 & 42.35 & \textbf{53.38} \\
\bottomrule
\end{tabular}
\caption{LLM-as-a-judge win rates}
\label{evaluation_scores_llm_judge_dimension}
\end{subtable}
\hfill
\begin{subtable}{0.49\textwidth}
\centering
\setlength{\tabcolsep}{0.2pt}
\renewcommand{\arraystretch}{0.875}
{\settowidth{\DeltaColW}{00.00}
\DeltaSetRange{0}{170}
\begin{tabular}{lcccccccc}
\toprule
\multirow{4.6}{*}{\textbf{model}} & \multicolumn{4}{c}{\textbf{Soundness}} & \multicolumn{4}{c}{\textbf{Overall assessment}} \\ \cmidrule(lr{3pt}){2-5} \cmidrule(lr{3pt}){6-9}
& \multicolumn{2}{c}{\textbf{$\Delta_{\leq0.5}$\%}} & \multicolumn{2}{c}{\textbf{$\Delta_{MAE}$\%}} & \multicolumn{2}{c}{\textbf{$\Delta_{\leq0.5}$\%}} & \multicolumn{2}{c}{\textbf{$\Delta_{MAE}$\%}} \\ \cmidrule(lr{3pt}){2-3} \cmidrule(lr{3pt}){4-5} \cmidrule(lr{3pt}){6-7} \cmidrule(lr{3pt}){8-9}
& \textbf{e2e} & \textbf{jge} & \textbf{e2e} & \textbf{jge} & \textbf{e2e} & \textbf{jge} & \textbf{e2e} & \textbf{jge} \\ \midrule
G-O-20B  & \DeltaHeat{76.85} & \DeltaHeat{48.25} & \DeltaHeat{147.13} & \DeltaHeat{104.33} & \DeltaHeat{87.13} & \DeltaHeat{56.31} & \DeltaHeat{150.87} & \DeltaHeat{101.34} \\
G-O-120B & \DeltaHeat{39.30} & \DeltaHeat{34.87} & \DeltaHeat{72.63} & \DeltaHeat{51.40} & \DeltaHeat{46.00} & \DeltaHeat{41.38} & \DeltaHeat{67.16} & \DeltaHeat{70.90} \\ \hdashline\noalign{\vskip 2pt}
Q3-1.7B   & \DeltaHeat{48.79} & \DeltaHeat{27.07} & \DeltaHeat{105.25} & \DeltaHeat{34.69} & \DeltaHeat{29.44} & \DeltaHeat{22.36} & \DeltaHeat{29.15}  & \DeltaHeat{27.50} \\
Q3-8B     & \DeltaHeat{46.55} & \DeltaHeat{2.31}  & \DeltaHeat{60.01}  & \DeltaHeat{10.20} & \DeltaHeat{49.07} & \DeltaHeat{14.80} & \DeltaHeat{78.75}  & \DeltaHeat{14.67} \\
Q3-14B    & \DeltaHeat{33.87} & \DeltaHeat{12.50} & \DeltaHeat{49.11}  & \DeltaHeat{13.95} & \DeltaHeat{54.11} & \DeltaHeat{19.38} & \DeltaHeat{117.02} & \DeltaHeat{15.85} \\
Q3-32B    & \DeltaHeat{48.87} & \DeltaHeat{10.00} & \DeltaHeat{81.44}  & \DeltaHeat{10.09} & \DeltaHeat{68.42} & \DeltaHeat{14.90} & \DeltaHeat{89.02}  & \DeltaHeat{15.75} \\ \hdashline\noalign{\vskip 2pt}
R1-Q-14B  & \DeltaHeat{14.28} & \DeltaHeat{16.19} & \DeltaHeat{29.77}  & \DeltaHeat{14.87} & \DeltaHeat{15.46} & \DeltaHeat{18.18} & \DeltaHeat{28.85}  & \DeltaHeat{16.86} \\
R1-Q-32B  & \DeltaHeat{22.87} & \DeltaHeat{18.77} & \DeltaHeat{36.02}  & \DeltaHeat{16.35} & \DeltaHeat{48.17} & \DeltaHeat{26.88} & \DeltaHeat{69.20}  & \DeltaHeat{27.13} \\ \hdashline\noalign{\vskip 2pt}
R1-L-8B  & \DeltaHeat{22.59} & \DeltaHeat{21.77} & \DeltaHeat{19.19}  & \DeltaHeat{15.08} & \DeltaHeat{8.61}  & \DeltaHeat{34.84} & \DeltaHeat{9.94}   & \DeltaHeat{27.43} \\
R1-L-70B & \DeltaHeat{28.69} & \DeltaHeat{14.08} & \DeltaHeat{47.88}  & \DeltaHeat{31.83} & \DeltaHeat{37.36} & \DeltaHeat{25.84} & \DeltaHeat{61.50}  & \DeltaHeat{26.01} \\ \midrule
\textsc{average} & 38.27 & \textbf{20.58} & 64.84 & \textbf{30.28} & 44.38 & \textbf{27.49} & 70.15 & \textbf{34.34} \\
\bottomrule
\end{tabular}}
\caption{Adversarial robustness}
\label{evaluation_scores_score_prediction_adversarial_part}
\end{subtable}
\caption{(a) Text similarity scores between end-to-end generations (\textbf{e2e}), judgment-grounded expansions (\textbf{jge}), and human-written reviews. (b) The percentage of score predictions within 0.5 points of the human-assigned scores (\textbf{$\leq$0.5}) and the mean absolute error (\textbf{MAE}). (c) LLM-as-a-judge win rates, TA = Technical Accuracy, CV = Constructive Value, AD = Analytical Depth. (d) Relative score changes under adversarial and non-adversarial settings. G-O = GPT-OSS, Q = Qwen, L = LLaMA, V = VL, R1 = Deepseek-R1, DeepR = DeepReview.}
\label{evaluation_scores_main}
\vspace{-1.6em}
\end{table*}

\subsubsection{The effect of reviewer judgment}
\label{value_of_judgment_grounded_expansion}

\setreturn{value_of_judgment_grounded_expansion}
We evaluate two dimensions to measure the effect of reviewer judgment. \textbf{Reference alignment} reflects how well a system follows the provided reviewer judgment by aligning with the reference review. We measure it through text similarity and recommendation accuracy for both content-level and overall stance alignment. \textbf{Practical utility} measures the practical value of judgment-grounded expansion. We assess it using LLM-as-a-judge, comparing generated reviews on several criteria \cite{zhu_deepreview_2025}, and adversarial robustness \cite{ye_are_2024}. We use the aspect-based proxy and generate one review comment per input to reduce confounds from variations in human input and the prediction paradigm. See \ref{more_on_gains_from_reviewer_judgment_signals} for more details.

For \textbf{reference alignment}, generated reviews under judgment-grounded expansion generally align better with the reference review. The relative gain over end-to-end generation in Table \ref{evaluation_scores_similarity} reflects how much additional text alignment is obtained when reviewer judgment is provided. We observe that larger models tend to have larger gains. Models trained with the DeepSeek-R1 distillation objective appear to benefit more from reviewer judgment, as shown by the comparison between base Qwen3 and R1-Distill-Qwen models. DeepReview also shows gains, indicating that a complex review generation system originally designed for end-to-end generation can still benefit from reviewer judgment. Table \ref{evaluation_scores_score_prediction} shows judgment-grounded expansion generally achieves higher \textbf{$\leq$0.5} and lower \textbf{MAE} scores, especially on \emph{Overall assessment}. This suggests that reviewer judgment helps anchor the generation to the reviewer's overall impression of a paper.

\setreturn{more_results}
Judgment-grounded expansion further shows \textbf{practical utility}. It achieves generally higher LLM-as-a-judge win rates on technical accuracy and analytical depth (see Table \ref{evaluation_scores_llm_judge_dimension}), possibly because reviewer judgment constrains the development of technical points and the scope of the generation, which reduces technically inaccurate expansion and makes the comment more focused. Table \ref{evaluation_scores_score_prediction_adversarial_part} shows both $\Delta_{\leq0.5}$\% and $\Delta_{MAE}$\% change substantially for end-to-end generation, while judgment-grounded expansion shows much smaller changes under the adversarial setting. Anchoring generation in reviewer judgment makes the system less affected by prompt injection, which is important for maintaining accountability.

We do not claim that judgment-grounded expansion universally outperforms end-to-end generation. Our results show that the reviewer judgment does not always lead to positive gains. This further highlights the need for reproducible large-scale evaluation to inspect how factors such as backbone models affect judgment-grounded expansion.

We report several complementary analyses in Appendix \ref{more_on_results}. We study the effect of quality check refinement and observe further gains in some cases (\ref{quality_check_refinement}). We manually analyze cases where judgment-grounded expansion underperforms the end-to-end baseline to characterize when reviewer judgment fails to help (\ref{manual_analysis}). We report results on ICLR and compare the exemplar-based and aspect-based proxies (\ref{other_results}).

\subsubsection{Judgment-utilization ability}
\label{collaborative_ability}

Building on our simulation and evaluation pipeline, we introduce \emph{judgment-utilization ability} as a diagnostic measure of the gains a system obtains from the reviewer judgment. This measure is not intended to be a setting-independent capability score. It is conditioned on the task, domain, metric, and model pool. For each model $m$ and condition $l$ (task, domain, metric), we denote by $p_{m,l}^{j}$ and $p_{m,l}^{e}$ its judgment-grounded expansion and end-to-end performance. We compute the normalized gain\footnote{Since metrics differ in direction, we negate lower-is-better metrics before normalization.} and then aggregate it to obtain a single \textbf{JUScore}:

{\small
\setlength{\abovedisplayskip}{2pt}
\setlength{\belowdisplayskip}{2pt}
\begin{equation}
\begin{aligned}
    \tilde p_{m,l}^{s}
    &= \frac{p_{m,l}^{s}}{\max_{m^{'} \in \mathcal{M}}(p_{m^{'},l}^{j}, p_{m^{'},l}^{e})}, s \in \{j, e\};\\[4pt]
    JUS&core(m) = \frac{1}{|\mathcal{L}|}\sum_{l}^{\mathcal{L}}(\tilde p_{m,l}^{j} - \tilde p_{m,l}^{e}),
\end{aligned}
\end{equation}
}

\noindent where $\mathcal{M}$ is the set of backbone models and $\mathcal{L}$ is the set of conditions. The JUScore provides a \uline{task}-, \uline{domain}-, and \uline{metric}-specific perspective of a model's ability to utilize reviewer judgment.

\begin{figure}[!ht]
\centering
\includegraphics[width=0.8\linewidth]{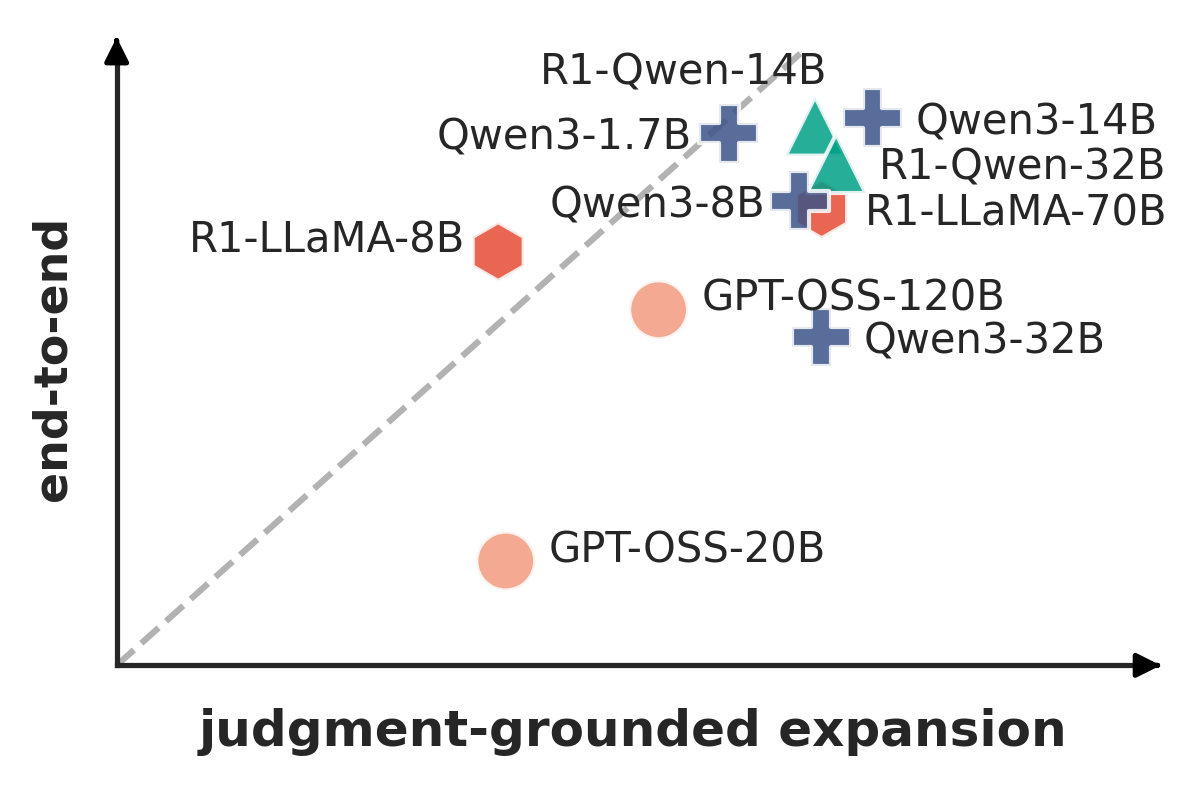}
\caption{End-to-end versus judgment-grounded expansion performance. The horizontal offset to the dashed line ($ y=x $) corresponds to a model's JUScore.}
\label{collaborability_scatter}
\vspace{-0.2\baselineskip}
\end{figure}

We compute JUScore on the review generation task for ARR papers using text similarity, recommendation accuracy, and adversarial robustness. Figure \ref{collaborability_scatter} visualizes the resulting JUScores.

In our setting, Qwen3-32B shows the largest gains from reviewer judgment (the largest positive offsets from $ y=x $). Higher end-to-end performance may not imply a high JUScore: the Pearson coefficient between end-to-end performance and JUScore is moderately negative but not statistically significant ($r=-0.5058$, $p=0.0816$). JUScore is a diagnostic that shows which models in a given pool benefit most from reviewer input. Its value is in informing backbone selection in a specific setting, not in producing a universal ranking.

\section{Balancing size and coverage}
\label{balancing_set_size_and_coverage}

Compared with single-point prediction, set prediction is a more practical setting for judgment-grounded expansion. By generating multiple possible expansions of a reviewer judgment, set prediction may increase the likelihood of covering the reviewer's intended comment. Though our user study participants found multiple candidates helpful, they commented that having multiple candidates adds cognitive load (see \S\ref{user_study}). This leads to a key practical question: how to curate a compact set while preserving coverage of the reviewer-selected review comment. Our user study data provides a useful resource for this analysis, as it records both the generated candidate set and the user's selected candidate for each input.

A straight forward way to curate a compact candidate set is \textbf{score-based ranking}, assigning each candidate a score and keeping the top-$k$ candidates. We evaluate 2 types of commonly used and interpretable scoring methods: (1) \emph{uncertainty-based} scores, including negative log-likelihood (NLL), maximum sequence probability (MSP), mean token entropy (MTE) \cite{fomicheva2020mte}, which are calculated using the candidate's generation probabilities; (2) \emph{similarity-based} score, calculated as SBERT similarity between the reviewer input and the candidate \cite{reimers2019sentencebert}.

\begin{table}[!ht]
\small
\centering
\setlength{\tabcolsep}{8pt}
\begin{tabular}{lcccc}
\toprule
\textbf{method} & \textbf{@1} & \textbf{@2} & \textbf{@3} & \textbf{avg. rank} \\ \midrule
\textit{random}  & 26.21 & 52.49 & 78.59 & 2.47 \\ \hdashline\noalign{\vskip 2pt}
NLL              & 33.82 & 58.82 & 80.88 & 2.28 \\
MSP              & \textbf{42.65} & 64.71 & 82.35 & 2.13 \\
MTE              & 41.18 & 63.24 & 82.35 & 2.16 \\
Sim              & 35.29 & \textbf{66.18} & 85.29 & 2.16 \\
MSP+Sim          & 41.18 & 64.71 & \textbf{86.76} & \textbf{2.09} \\
\bottomrule
\end{tabular}
\caption{Score-based ranking results. We report the percentage of cases where the human-selected candidate ranks first (\textbf{@1}) or within the top two or three (\textbf{@2}, \textbf{@3}), and its average rank (\textbf{avg. rank}).}
\label{score_based_curation}
\end{table}

Table \ref{score_based_curation} shows the results of score-based ranking. Results are based on 68 interactions where users selected a candidate, with an average of 3.93 candidates per instance. Compared with the random baseline, both uncertainty-based and similarity-based scores capture meaningful signals related to user preference. We further combine the strongest \textbf{@1} and \textbf{@2} methods, i.e., MSP+Sim, by normalizing MSP and Sim separately within the candidate set and summing the normalized scores. The combined score improves \textbf{@3} and yields the best average rank, suggesting that different methods may provide complementary signals.  

A limitation of score-based set curation is that it produces prediction sets of fixed size. In practice, however, the desired set size should adapt to instance-level ``signals'' such as uncertainty or confidence. Conformal prediction is therefore better suited to this setting: by calibrating a threshold on a held-out calibration set, it returns variable-size prediction sets with a formal guarantee of target coverage \cite{angelopoulos2024conformal}.

We split the 68 interactions where users selected a candidate evenly into calibration and test sets. To reduce variance from a single split, we report average results over 5 random splits. Each calibration instance consists of a user input $x$, a generated candidate set $\mathcal{Y}(x)$, and the human-selected candidate $c^\ast$. For a target coverage $1-\alpha$, conformal prediction estimates a threshold $\tau_\alpha$ from the nonconformity scores of human-selected candidates in the calibration set. At test time, it returns all candidates $\mathcal{C}_{\alpha}(x)$ whose nonconformity scores are below this threshold:

\begin{equation}
    \mathcal{C}_{\alpha}(x) = \{c:S(x,c)\leq \tau_\alpha\}.
\end{equation}

\noindent Conformal prediction constructs an adaptive candidate set by applying this calibrated threshold at test time. We evaluate two types of nonconformity scores. For \emph{raw-value} nonconformity, $S(x,c)$ is computed directly from the scoring function. For \emph{rank-based} nonconformity, $S(x,c)$ is the candidate’s rank within $\mathcal{Y}(x)$ according to the scoring function \cite{huang2024conformalrank}. We evaluate the same scoring functions as in score-based ranking. We exclude NLL and MSP from \emph{raw-value} nonconformity, since their absolute values are not comparable across inference instances.

\begin{figure}[!t]
\centering
\begin{subfigure}[b]{0.92\linewidth}
\centering
\includegraphics[width=\linewidth]{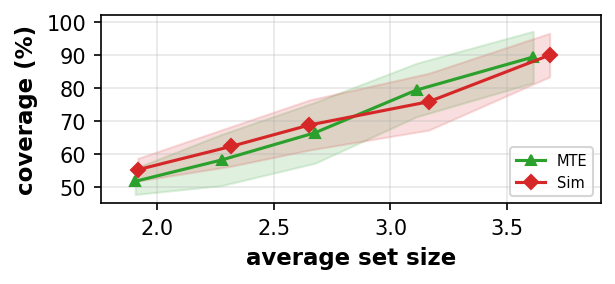}
\caption{\emph{raw-value}}
\label{fig:sub1}
\end{subfigure}

\begin{subfigure}[b]{0.92\linewidth}
\centering
\includegraphics[width=\linewidth]{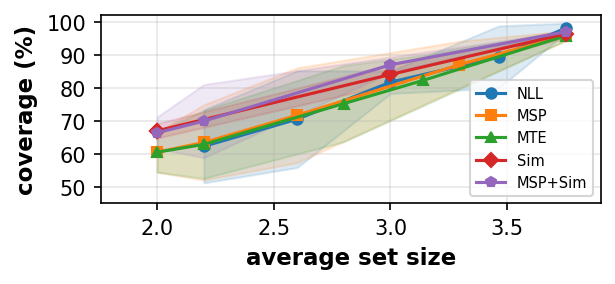}
\caption{\emph{rank-based}}
\label{fig:sub2}
\end{subfigure}
\caption{Conformal prediction results. We test $\alpha \in \{0.1,0.2,0.3,0.4,0.5\}$. Each point indicates the coverage rate of the human-selected candidate and the average candidate set size. Shaded regions indicate variation across random calibration/test splits. The target coverage is $1-\alpha$, but empirical coverage may not exactly match it due to finite calibration/test samples.}
\label{cp_based_curation}
\end{figure}

Figure \ref{cp_based_curation} shows that \emph{rank-based} nonconformity generally provides a better size and coverage trade-off than \emph{raw-value} nonconformity, achieving higher coverage at similar average set sizes. Among \emph{rank-based} methods, Sim and MSP+Sim perform particularly well. Compared with fixed score-based ranking, conformal prediction better supports judgment-grounded expansion: it allows the system to adjust candidate set size according to a nonconformity score, and by selecting different target coverage levels, practitioners can manually tune whether the system favors broader candidate sets with higher coverage or smaller sets that reduce reviewer cognitive load. This makes conformal prediction a promising technique for balancing target coverage against reviewer cognitive load.

Though the scoring functions capture reviewer preference to some extent, there remains room for improvement. Future work could design nonconformity scores tailored to peer review by incorporating review-specific factors, such as actionability, verifiability, and helpfulness \cite{sadallah2025good}.




\section{Conclusion}

Judgment-grounded expansion is more than an alternative for producing reviews. It accommodates a genuine need that balances automation and accountability. Our work presents a systematic study of judgment-grounded expansion, establishes it as a concrete collaboration mode, and provides a foundation for future review generation systems.

\clearpage

\section*{Limitations}

\noindent\textbf{Reviewer expertise.}\ \ Judgment-grounded expansion relies on the reviewer’s ability to provide reasonable evaluative claims and assess generated comments. As a result, this setting is potentially less suitable for less experienced reviewers.

\vspace{0.3em}

\noindent\textbf{Simulation.}\ \ (1) Our simulation does not aim to capture human preferences. Our goal is therefore not to replace human evaluation, but to complement it by enabling reproducible large-scale evaluation for analyzing task and model factors. (2) For human input simulation, the exemplar-based proxy captures broad patterns of real user input but may be biased toward the most common input style. Our aspect-based proxy relies on the aspect tagger from \citet{lu_identifying_2025}, which is not perfectly accurate. This is also a reason why we treat the aspect-based proxy as a low-information proxy: it may provide an inaccurate signal about the reviewer's intended key point. As a result, findings based on the aspect-based proxy, as well as quality check results involving aspect information, may be affected by tagging errors. Since more accurate aspect labels would yield stronger evaluative signals, this limitation likely makes our estimates conservative. (3) For quality check simulation, the \emph{Specification} check shows lower human-model and human-human agreement, suggesting that simulating this dimension remains an open question for future work. (4) Since it is impractical to compare simulated results with corresponding human-involved results, we inspect the validity of our simulation through sanity checks on its two key components, i.e., human input and quality check. While these checks provide supporting evidence, they do not constitute a full validation. In particular, the human evaluation used to inspect the simulated quality check is limited in scale.

\vspace{0.3em}

\noindent\textbf{User study.}\ \ The scale of our user study is limited, which may affect the generality of the findings based on this data in \S\ref{user_study} and \S\ref{balancing_set_size_and_coverage}. However, scaling such studies is challenging as participants must read papers, form judgments, and interact with the system, which is time-consuming. Despite this limitation, our study provides an initial resource for studying judgment-grounded expansion.

\vspace{0.3em}

\noindent\textbf{Experimental setup.}\ \ (1) We do not experiment with closed-source backbone models. Since our target is formal settings such as conference reviewing, we consider open-source models more appropriate, as they better support transparency and deployability. (2) Since DeepReview is a pre-configured system and we could not modify its generation pipeline, we append the judgment-grounded expansion prompt to the paper text as input. This is not a fair setup for DeepReview. Nevertheless, we still observe gains from reviewer judgment in this case. (3) We do not have access to pre-rebuttal review scores in both the ARR and ICLR data, which may introduce a version mismatch between the review scores and the review text. Thus, we report the percentage of score predictions within 0.5 points of the human-assigned scores in our evaluation to partially mitigate this issue.

\vspace{0.3em}

\noindent\textbf{Evaluation metrics.}\ \ The evaluation methods we use are not direct measures of review quality, and each metric has its own shortcomings. We adopt them for two reasons. First, they follow common practice in prior work. Second, they capture different perspectives, which together provide a more well-rounded view of model performance.

\vspace{0.3em}

\noindent\textbf{Reproducibility.}\ \ The reproducibility of quality check simulations using OpenAI o3-mini may be limited, as it is a closed-source model. We chose this model because quality check simulation is a key component of our pipeline that may benefit from strong reasoning ability. To mitigate this, we release all prompts and the simulation outputs.

\section*{Ethics statement}

Judgment-grounded expansion mitigates some accountability concerns of fully automatic review generation by anchoring the content in the reviewer's own evaluative claim. However, it does not resolve questions regarding authorship, disclosure, and responsibility for the review. Our work explores how to better balance automation and accountability, rather than offering a complete solution to these questions.

Our experiments use publicly available peer review data licensed for research use. The user study was conducted with ethics approval.

We used an AI assistant to improve the writing and language of the manuscript.

\section*{Acknowledgments}

Funded/Co-funded by the European Union (ERC, InterText, 101054961). Views and opinions expressed are however those of the author(s) only and do not necessarily reflect those of the European Union or the European Research Council. Neither the European Union nor the granting authority can be held responsible for them. This research work has been funded by the German Federal Ministry of Research, Technology and Space and the Hessian Ministry of Higher Education, Research, Science and the Arts within their joint support of the National Research Center for Applied Cybersecurity ATHENE. We gratefully acknowledge support from the hessian.AI Service Center (funded by the Federal Ministry of Research, Technology and Space, BMFTR, grant no. 16IS22091) and the hessian.AI Innovation Lab (funded by the Hessian Ministry for Digital Strategy and Innovation, grant no. S-DIW04/0013/003).

\bibliography{custom}

\clearpage

\appendix

\section{More on related work}

Table \ref{evaluation_methods_related_work} summarizes common evaluation metrics used in prior work for review evaluation.

\begin{table}[!ht]
\scriptsize
\centering
\begin{tabular}{>{\raggedright\arraybackslash}m{0.24\linewidth}m{0.56\linewidth}}
\toprule
\textbf{method} & \textbf{related work} \\ \midrule
Text similarity & \citet{chamoun_automated_2024,faizullah_limgen_2024,gao_reviewer2_2024,tan_peer_2024,zhou_is_2024,gao_reviewagents_2025} \\ \midrule
Recommendation & \citet{yuan_kid-review_2022,lu_ai_2024,zhou_is_2024,idahl_openreviewer_2025,zhu_deepreview_2025,weng_cycleresearcher_2024} \\ \midrule
LLM-as-a-judge & \citet{gao_reviewer2_2024,zhou_is_2024,idahl_openreviewer_2025,zhu_deepreview_2025} \\ \midrule
Human evaluation & \citet{wang_reviewrobot_2020,yuan_kid-review_2022,liang_can_2024,liu_reviewergpt_2023,chamoun_automated_2024,du_llms_2024,tyser_ai-driven_2024,zhou_is_2024} \\
\bottomrule
\end{tabular}
\caption{Common evaluation methods used in literature. \backtomain{review_evaluation}}
\label{evaluation_methods_related_work}
\end{table}

\section{More on user study}
\label{more_on_user_study}

We implement a web-based interface for judgment-grounded expansion. The interface has two main components, a PDF reader on one side, and an interaction input panel on the other where users enter evaluative claims and inspect system outputs. Users initiate the process by typing an evaluative claim (e.g., ``The novelty is clear'') into the input panel. The system then generates a set of review comment candidates using a two-step pipeline: it first prompts an LLM to identify 2-5 plausible angles implied by the claim with respect to the paper, and then generates one candidate comment per angle in a separate call, conditioned on the paper, the claim, and the corresponding angle. We adopt this two-step generation pipeline so that candidates are meaningfully different along interpretable dimensions, rather than paraphrases of each other. We use GPT-OSS-120B as the backbone model. This choice was made before our experiments and was not based on the findings reported in the paper. We selected it as a strong large open-source model that we expected to perform well. For each inference, we record all the generated candidates and their \texttt{logprobs}. See \href{https://anonymous.4open.science/r/judgment-grounded-expansion-B28D/user_study}{here} for the implementation of the interface.

We recruited 6 participants through an open call. All participants are PhD students with at least 2 years of research experience. English is their primary language for reading research papers, and all participants have prior peer review experience. Participants were compensated with vouchers. No personal identifiable information was collected.

\setreturn{more_on_simulation}
Participants completed a consent form describing the purpose of the study, the artifacts being collected, privacy considerations, and the intended use of the data. Participants completed a background questionnaire indicating their areas of expertise. We assigned papers based on their responses to ensure a reasonable match between participant expertise and paper topic. Participants then completed a familiarity check to confirm that they were neither overly familiar with the assigned paper nor entirely unfamiliar with the topic. All papers were anonymized initial drafts selected from \href{https://tudatalib.ulb.tu-darmstadt.de/handle/tudatalib/4460}{ARR}.

In total, participants completed reviews for 14 papers, with each participant completing 2-3 reviews. There is an average of 5.8 user-system interactions for each paper. 

After completing each review, participants answered the following 7-point Likert-scale questions: (1) Post-edit effort: The effort I needed to rewrite or correct the generated review before submitting was. (2) Confidence: I am confident that the final review reflects the key points I intended to raise. (3) Satisfaction: I am satisfied with the final review. (4) Preference: Compared to writing a review from scratch, I prefer using judgment expansion. Beyond the results reported in the main text, participants also reported moderately high confidence in the final reviews (Mean=5.54).

After the full study, participants answered the following open-ended questions: (1) What was the most helpful aspect of the judgment-grounded expansion workflow, and why? (2) What was the most frustrating aspect of the judgment-grounded expansion workflow, and why? (3) With the judgment-grounded expansion workflow, do you feel responsible for the final review, and why? (4) Compared to writing from-scratch, do you feel that judgment-grounded expansion allowed you to produce a review more efficiently, and why? (5) How did the set of multiple candidates affect your experience, did having options help or add cognitive burden? Apart from the findings reported in the main text, participants identified several limitations of the workflow. They noted that generated comments were sometimes overly verbose, refinement was occasionally insufficient, and the interface did not allow them to select and combine multiple candidates. These issues point to system design improvements for future judgment-grounded expansion tools. Participants also reported feeling responsible for the final review, largely because the review was grounded in their own judgments. \backtomain{user_study}

\setreturn{more_on_experimental_settings}

\section{More on simulation}
\label{more_on_simulating_collaborative_generation}

Table \ref{examples_of_aspects} shows examples of \textsc{coarse} and \textsc{fine} aspects and the review comments from which they are extracted. Table \ref{creating_exemplar_based_proxy} shows the prompt for creating exemplar-based proxies. Table \ref{quality_check_prompts} shows the prompts for the \emph{Evidence} and \emph{Reasoning} checks.

\begin{table}[!ht]
\scriptsize
\centering
\begin{tabular}{m{0.43\linewidth}m{0.2\linewidth}m{0.2\linewidth}}
\toprule
\textbf{review comment} & \textbf{\textsc{coarse}} & \textbf{\textsc{fine}} \\ \midrule
The proposed method demonstrates commendable innovation, standing apart from mere amalgamation of existing models. & Methodology, Novelty & Method,\newline Novelty \\ \midrule
If the proposed model introduces high computational complexity and yields only marginal gains, its real-world utility may be limited. & Methodology & Complexity \\ \midrule
The paper lacks comparative analysis with some important baselines, such as P-tuning v2. & Analysis,\newline Comparison & Analysis,\newline Baseline \\ \midrule
The exclusive use of a single language model, T5-base, as the backbone prompts doubts about the general applicability of the proposed approach across a broader spectrum of models. & Methodology & Application, Model \\
\bottomrule
\end{tabular}
\caption{Examples of aspects and the review comments from which they are extracted \cite{lu_identifying_2025}. \backtomain{simulating_human_input}}
\label{examples_of_aspects}
\end{table}

\section{More on experimental settings}
\label{more_on_experiment_settings}

We perform at most two rounds of generation for each quality check dimension. For example, if a generation fails the \emph{Relevance} check, it is re-generated once. If the new generation still fails the check, the generation process stops.

Tables \ref{end_to_end_prompt}, \ref{collaborative_prompt}, and \ref{collaborative_prompts_next_round} show the review generation prompts used in our study. Our prompts incorporate guidelines from the official \href{https://web.archive.org/web/20240421045330/https://aclrollingreview.org/reviewform}{ARR review form}. A web cache is provided because the ARR review form for ARR is different from the current version. The initial generation for judgment-grounded expansion is performed in a single step (see Table \ref{collaborative_prompt}). In the refinement step, generation is performed on a per-key-point basis, using a chat-style setup that carries over all inputs and outputs from the previous round (see Table \ref{collaborative_prompts_next_round}).

We experiment with a range of models to generate reviews: GPT-OSS (20B, 120B) \cite{openai_gpt-oss-120b_2025}, Qwen3 (1.7B, 8B, 14B, 32B), Qwen3-VL-Instruct (8B, 32B) \cite{yang_qwen3_2025}, R1-Distill-Qwen (14B, 32B), and R1-Distill-LLaMA (8B, 70B) \cite{deepseek-ai_deepseek-r1_2025}.

For text-only models, we convert paper draft PDFs into plain text using \href{https://pypdf2.readthedocs.io/en/3.x/}{PyPDF2}. For vision-language models, we use \href{https://pymupdf.readthedocs.io/en/latest/index.html}{PyMuPDF} to convert PDFs into PNG images.

Experiments using models except GPT-OSS and o3-mini were conducted locally using \href{https://huggingface.co/}{\texttt{HuggingFace Transformers}}. All experiments using GPT-OSS models were conducted locally using vLLM \cite{kwon_efficient_2023} (we did not use HuggingFace Transformers due to a memory leakage issue with GPT-OSS models). Inferences with o3-mini were performed via the OpenAI API. For all local inferences, we set \texttt{temperature=0.8}, \texttt{max\_tokens=2048}, and \texttt{seed=2266}. For o3-mini, we set \texttt{verbosity=``low''}. We use \texttt{google/bigbird-roberta-base} for BERTScore, as it supports long input sequences (up to 4096 tokens) \cite{zaheer_big_2020}. All local experiments were conducted on NVIDIA A100 80GB GPUs.

We use DeepReviewer-14B with the generation mode set to \texttt{standard} \cite{zhu_deepreview_2025}. Since we could not modify its generation pipeline to generate reviews on a key-point-and-judgment basis, we append the judgment-grounded expansion prompt to the paper text as input. While this is not the optimal configuration, it is the most feasible setup given the constraints. Given this, we only conduct a limited number of experiments with this system. \backtomain{review_generation}

\setreturn{more_on_recommendation_accuracy}
\section{More on results}
\label{more_on_results}

\subsection{More on sanity check}

We conduct human evaluation to assess the alignment between our simulated quality checks and human assessments. We evaluate 100 queries, which correspond to reviews from 17 papers. Each item is independently assessed by 2 human evaluators. For each query, evaluators first read the paper, and then assess whether a candidate review segment aligns with the intent of a reference review segment, i.e., performing quality check across \emph{Relevance}, \emph{Specification}, \emph{Evidence}, and \emph{Reasoning}. To reduce possible bias, evaluators were not told that the candidate review is LLM-generated and the reference is human-written. Both evaluators are Computer Science graduate students with prior peer review experience.

\begin{table}[!ht]
\centering
\small
\setlength{\tabcolsep}{3.7pt}
\begin{tabular}{lccc ccc ccc}
\toprule
\multirow{2.5}{*}{\textbf{dim.}} & \multicolumn{3}{c}{\textbf{H1-Model}} & \multicolumn{3}{c}{\textbf{H2-Model}} & \multicolumn{3}{c}{\textbf{H1-H2}} \\
\cmidrule(lr){2-4}\cmidrule(lr){5-7}\cmidrule(lr){8-10}
& $P_o$ & AC\textsubscript{1} & $n$ & $P_o$ & AC\textsubscript{1} & $n$ & $P_o$ & AC\textsubscript{1} & $n$ \\
\midrule
Rele & 0.76 & \uline{0.69} & 82 & 0.60 & 0.34 & 82 & 0.72 & \uline{0.63} & 82 \\
Spec & 0.46 & -0.07 & 74 & 0.52 & 0.11 & 52 & 0.63 & 0.35 & 59 \\
Evi  & 0.72 & \uline{0.57} & 40 & 0.72 & \uline{0.51} & 43 & 0.83 & \uline{0.67} & 35 \\
Reas & 0.95 & \uline{0.94} & 38 & 0.97 & \uline{0.97} & 39 & 0.97 & \uline{0.97} & 35 \\
\bottomrule
\end{tabular}
\caption{Inter-annotator quality check agreement between two human evaluators (\textbf{H1} and \textbf{H2}) and o3-mini (\textbf{Model}). We report percent agreement ($P_o$), Gwet's AC\textsubscript{1} (AC\textsubscript{1}), and the number of overlapping labeled items ($n$). \uline{Underlined} AC\textsubscript{1} values indicate AC\textsubscript{1} $>$ 0.5, corresponding to moderate-to-good agreement. \backtomain{sanity_check}}
\label{iaa_quality_check}
\end{table}

Table \ref{iaa_quality_check} shows the inter-annotator agreement for the quality check, including both human-model and human-human agreement. We report (1) percent agreement, the fraction of items receiving identical labels, and (2) Gwet’s AC\textsubscript{1} \cite{gwet_ac1_2008}, an agreement coefficient that is more robust under imbalanced label distributions.

\begin{figure}[!t]
\centering
\begin{subfigure}[b]{\linewidth}
    \centering
    \includegraphics[width=0.9\linewidth]{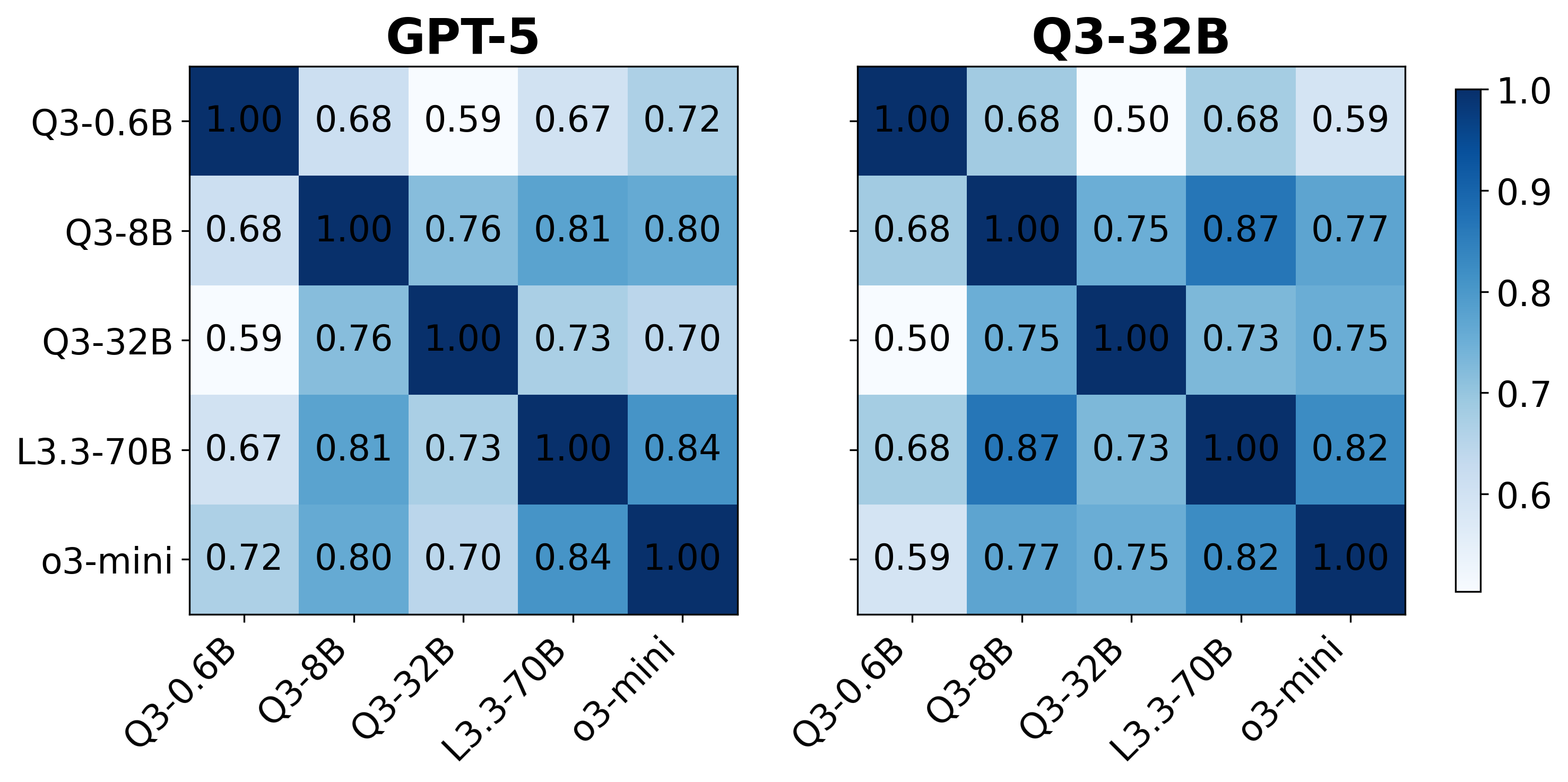}
    \caption{Evidence}
    \label{quality_check_consistency_evidence_modified}
\end{subfigure}

\vspace{0.2em}

\begin{subfigure}[b]{\linewidth}
    \centering
    \includegraphics[width=0.9\linewidth]{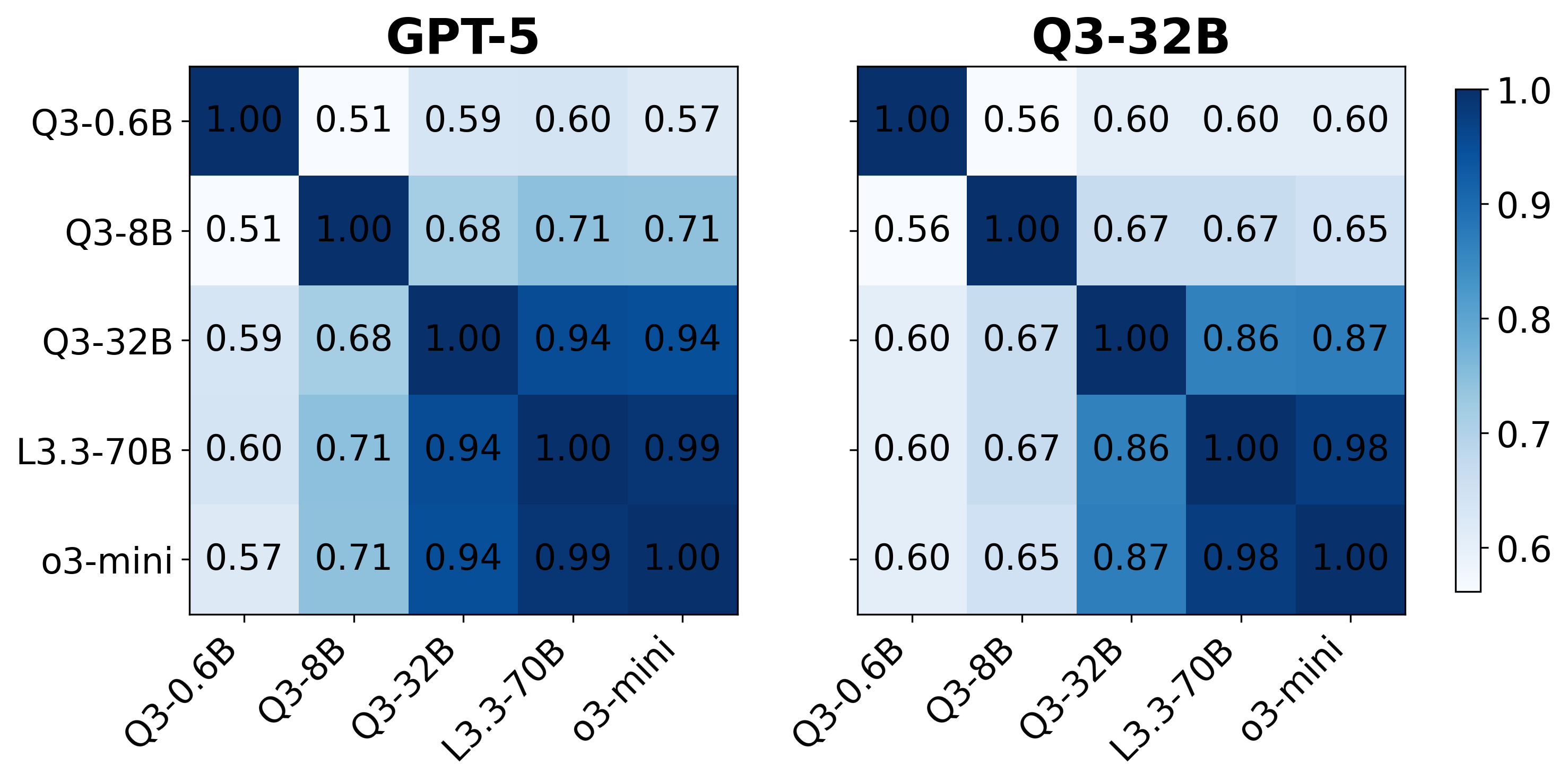}
    \caption{Reasoning}
    \label{quality_check_consistency_reasoning}
\end{subfigure}
\caption{Quality check consistency across models. Each subfigure title indicates the model used for review generation, and the x- and y-axes indicate the models used for quality check. Q = Qwen, L = LLaMA. \backtomain{sanity_check}}
\label{quality_check_consistency}
\end{figure}

Figure \ref{quality_check_consistency} shows the consistency of the quality check simulation across different LLMs.

\setreturn{more_on_gains_from_reviewer_judgment_signals}
\subsection{More on the effect of reviewer judgment}
\label{more_on_gains_from_reviewer_judgment_signals}

We assemble review segments into complete reviews when comparing with end-to-end generation or performing review score prediction. In some evaluations, we assemble reviews using Pass-or-First (PoF) segments: we use the final round segment only if it passes all quality checks, otherwise fall back to the first round segment.
We evaluate recommendation accuracy by predicting review scores and comparing them against the human-assigned scores. We report the percentage of predictions within 0.5 points of the human-assigned scores (\textbf{$\leq$0.5}) and the mean absolute error (\textbf{MAE}). We prefer \textbf{$\leq$0.5} over exact match, as it allows minor discrepancies that do not substantially alter the overall stance, providing a more faithful view of the overall performance. For ARR, we targeted \emph{Soundness} and \emph{Overall assessment}. For ICLR 2025, we targeted \emph{Soundness}, \emph{Presentation}, \emph{Contribution}, and \emph{Rating}. The prompts are shown in Tables \ref{score_prediction_prompt_arr} and \ref{score_prediction_prompt_iclr}.

For LLM-as-a-judge, we follow the prompt from \citet{zhu_deepreview_2025}, evaluating technical accuracy, constructive value, and analytical depth (see Table \ref{llm_judge_prompt}). To mitigate positional bias commonly observed in LLM-as-a-judge evaluations, we adopted the Balance Position Calibration (BPC) strategy \cite{liu_lost_2024,wang_large_2024}, performing inference twice: once with end-to-end generation as the first candidate and once with collaborative generation as the first. We used LLaMA3.3-70B-Instruct as the judge model \cite{grattafiori_llama_2024}.

We assess adversarial robustness by measuring the change in review score predictions between non-adversarial and adversarial settings, where smaller changes indicate greater robustness. We inject a prompt into the paper text (see Table \ref{adversarial_text}).

\subsubsection{Quality check refinement}
\label{quality_check_refinement}

Table \ref{quality_check_pass_counts} reports the raw counts of quality check passes across models and generation rounds.

\begin{table}[!ht]
\scriptsize
\centering
\begin{tabular}{lccccc}
\toprule
\textbf{model} & \textbf{round} & \textbf{Rele} & \textbf{Spec} & \textbf{Evi} & \textbf{Reas} \\ \midrule
\multirow{2}{*}{G-O-20B} & gen & 323/562 & 90/323 & 58/90 & 57/58 \\
& refin & 58/224 & 131/230 & 14/27 & 1/1 \\ \midrule
\multirow{2}{*}{G-O-120B} & gen & 339/585 & 99/339 & 62/99 & 62/62\\
& refin & 114/243 & 147/239 & 24/35 & 0/0 \\ \midrule
\multirow{2}{*}{Q3-1.7B} & gen & 190/353 & 57/190 & 25/57 & 24/25 \\
& refin & 32/92 & 32/98 & 11/22 & 1/1 \\ \midrule
\multirow{2}{*}{Q3-8B} & gen & 294/505 & 89/294 & 51/89 & 51/51 \\
& refin & 40/211 & 106/204 & 11/36 & 0/0 \\ \midrule
\multirow{2}{*}{Q3-14B} & gen & 338/505 & 103/338 & 62/103 & 62/62 \\
& refin & 45/166 & 129/235 & 18/38 & 0/0 \\ \midrule
\multirow{2}{*}{Q3-32B} & gen & 376/540 & 120/376 & 79/120 & 78/79 \\
& refin & 48/164 & 161/256 & 10/38 & 1/1 \\ \midrule
\multirow{2}{*}{Q3-VL-8B} & gen & 255/457 & 84/255 & 51/84 & 50/51 \\
& refin & 53/199 & 86/171 & 7/27 & 1/1 \\ \midrule
\multirow{2}{*}{Q3-VL-32B} & gen & 441/590 & 131/441 & 84/131 & 83/84 \\
& refin & 54/149 & 185/310 & 19/47 & 1/1 \\ \midrule
\multirow{2}{*}{R1-Q-14B} & gen & 259/455 & 86/259 & 49/86 & 47/49 \\
& refin & 45/187 & 73/151 & 9/35 & 2/2 \\ \midrule
\multirow{2}{*}{R1-Q-32B} & gen & 369/541 & 115/369 & 75/115 & 75/75 \\
& refin & 57/165 & 111/210 & 16/36 & 0/0 \\ \midrule
\multirow{2}{*}{R1-L-8B} & gen & 94/259 & 25/94 & 15/25 & 13/15 \\
& refin & 24/117 & 13/41 & 2/10 & 2/2 \\ \midrule
\multirow{2}{*}{R1-L-70B} & gen & 332/519 & 105/332 & 68/105 & 68/68 \\
& refin & 56/147 & 93/183 & 17/37 & 0/0 \\
\bottomrule
\end{tabular}
\caption{Pass counts for each quality check dimension across models and rounds. We report the number of passes over the total number of cases. G = GPT, O = OSS, Q = Qwen, L = LLaMA.}
\label{quality_check_pass_counts}
\end{table}

Table \ref{quality_check_pass_and_not_pass} shows text similarity scores between first round generations that pass or fail the quality checks. For R1-Distill models, generations that pass a given check generally achieve higher scores than those that fail the same check, suggesting that the quality checks provide meaningful signals regarding the similarity between the generations and human-written reviews for these models. Across all models, generations that pass the \emph{Evidence} check achieve higher scores than those failing it. A plausible reason is that both human reviewers and LLMs tend to reference paper-specific contents, which increases lexical/semantic overlap.

{\setlength{\tabcolsep}{3pt}
\begin{table}[!t]
\scriptsize
\begin{subtable}{\linewidth}
\centering
\begin{tabular}{lcccccccc}
\toprule
\multirow{2.4}{*}{\textbf{model}} & \multicolumn{2}{c}{\textbf{Relevance}} & \multicolumn{2}{c}{\textbf{Specification}} & \multicolumn{2}{c}{\textbf{Evidence}} & \multicolumn{2}{c}{\textbf{Reasoning}} \\ \cmidrule(lr){2-3} \cmidrule(lr){4-5} \cmidrule(lr){6-7} \cmidrule(lr){8-9}
& \textbf{False} & \textbf{True} & \textbf{False} & \textbf{True} & \textbf{False} & \textbf{True} & \textbf{False} & \textbf{True} \\ \midrule
G-O-20B & 1.40 & \textbf{1.65} & 1.51 & \textbf{1.98} & 0.00 & \textbf{2.45} & \textbf{19.66} & 2.00 \\
G-O-120B & \textbf{1.33} & 1.27 & \textbf{1.33} & 1.04 & 0.00 & \textbf{1.31} & - & 1.31 \\ \hdashline\noalign{\vskip 2pt}
Q3-1.7B & 1.10 & \textbf{1.69} & \textbf{1.74} & 1.55 & 0.00 & \textbf{2.82} & 0.00 & \textbf{2.89} \\
Q3-8B & \textbf{2.58} & 2.19 & 1.92 & \textbf{2.76} & 0.92 & \textbf{3.72} & - & 3.72 \\
Q3-14B & \textbf{2.54} & 1.99 & 1.90 & \textbf{2.19} & 1.04 & \textbf{2.63} & - & 2.63 \\
Q3-32B & 2.34 & \textbf{2.59} & 2.55 & \textbf{2.65} & 0.00 & \textbf{3.42} & 0.00 & 3.42 \\ \hdashline\noalign{\vskip 2pt}
Q3-VL-8B    & \textbf{1.86} & 1.56 & \textbf{1.64} & 1.26 & 1.00 & \textbf{1.21} & 0.00 & \textbf{1.19} \\
Q3-VL-32B   & 1.33 & \textbf{1.75} & 1.72 & \textbf{1.84} & 1.27 & \textbf{2.04} & 0.00 & \textbf{2.06} \\ \hdashline\noalign{\vskip 2pt}
R1-Q-14B & \textbf{2.75} & 2.20 & 2.15 & \textbf{2.29} & 0.00 & \textbf{3.12} & 0.00 & \textbf{3.18} \\
R1-Q-32B & 2.05 & \textbf{2.64} & 2.34 & \textbf{3.28} & 0.98 & \textbf{4.05} & - & 4.05 \\ \hdashline\noalign{\vskip 2pt}
R1-L-8B & 0.97 & \textbf{2.40} & 2.01 & \textbf{3.54} & 0.00 & \textbf{4.91} & 0.00 & \textbf{5.04} \\
R1-L-70B & 2.16 & \textbf{2.74} & 2.51 & \textbf{3.23} & 0.00 & \textbf{3.56} & - & 3.56 \\
\bottomrule
\end{tabular}
\caption{BLEU $\uparrow$}
\end{subtable}

\vspace{1em}

\begin{subtable}{\linewidth}
\centering
\begin{tabular}{lcccccccc}
\toprule
\multirow{2.4}{*}{\textbf{model}} & \multicolumn{2}{c}{\textbf{Relevance}} & \multicolumn{2}{c}{\textbf{Specification}} & \multicolumn{2}{c}{\textbf{Evidence}} & \multicolumn{2}{c}{\textbf{Reasoning}} \\ \cmidrule(lr){2-3} \cmidrule(lr){4-5} \cmidrule(lr){6-7} \cmidrule(lr){8-9}
& \textbf{False} & \textbf{True} & \textbf{False} & \textbf{True} & \textbf{False} & \textbf{True} & \textbf{False} & \textbf{True} \\ \midrule
G-O-20B & 14.28 & \textbf{14.42} & 14.33 & \textbf{14.54} & 9.67 & \textbf{17.14} & \textbf{30.56} & 16.90 \\
G-O-120B & 14.61 & \textbf{14.98} & 14.50 & \textbf{16.09} & 11.99 & \textbf{18.55} & - & 18.55 \\ \hdashline\noalign{\vskip 2pt}
Q3-1.7B & 15.58 & \textbf{16.15} & 15.97 & \textbf{16.54} & 15.09 & \textbf{19.64} & 14.39 & \textbf{19.45} \\
Q3-8B & \textbf{16.63} & 16.29 & 16.20 & \textbf{16.46} & 11.73 & \textbf{20.04} & - & 20.04 \\
Q3-14B & \textbf{16.14} & 15.83 & 15.76 & \textbf{16.03} & 12.23 & \textbf{18.47} & - & 18.47 \\
Q3-32B & 15.80 & \textbf{16.81} & 16.37 & \textbf{17.68} & 15.60 & \textbf{20.83} & 11.88 & \textbf{20.73} \\ \hdashline\noalign{\vskip 2pt}
Q3-VL-8B    & \textbf{15.52} & 14.99 & 14.95 & \textbf{15.13} & 12.27 & \textbf{17.04} & 15.38 & \textbf{17.12} \\
Q3-VL-32B   & 14.44 & \textbf{14.48} & \textbf{14.64} & 14.11 & 11.23 & \textbf{15.72} & 6.90 & \textbf{15.83} \\ \hdashline\noalign{\vskip 2pt}
R1-Q-14B & 16.53 & \textbf{16.29} & 15.66 & \textbf{17.58} & 13.14 & \textbf{20.97} & 14.00 & \textbf{21.20} \\
R1-Q-32B & 15.13 & \textbf{16.17} & 15.57 & \textbf{17.53} & 11.96 & \textbf{20.51} & - & 20.51 \\ \hdashline\noalign{\vskip 2pt}
R1-L-8B & 13.20 & \textbf{17.14} & 17.02 & \textbf{17.29} & 11.90 & \textbf{20.95} & 20.06 & \textbf{21.16} \\
R1-L-70B & 16.06 & \textbf{16.76} & 16.29 & \textbf{17.74} & 0.00 & \textbf{20.04} & - & 20.04 \\
\bottomrule
\end{tabular}
\caption{ROUGE-L $\uparrow$}
\end{subtable}

\vspace{1em}

\begin{subtable}{\linewidth}
\centering
\begin{tabular}{lcccccccc}
\toprule
\multirow{2.4}{*}{\textbf{model}} & \multicolumn{2}{c}{\textbf{Relevance}} & \multicolumn{2}{c}{\textbf{Specification}} & \multicolumn{2}{c}{\textbf{Evidence}} & \multicolumn{2}{c}{\textbf{Reasoning}} \\ \cmidrule(lr){2-3} \cmidrule(lr){4-5} \cmidrule(lr){6-7} \cmidrule(lr){8-9}
& \textbf{False} & \textbf{True} & \textbf{False} & \textbf{True} & \textbf{False} & \textbf{True} & \textbf{False} & \textbf{True} \\ \midrule
G-O-20B & 60.26 & \textbf{60.45} & \textbf{60.55} & 60.20 & 56.05 & \textbf{62.48} & \textbf{68.58} & 62.38 \\
G-O-120B & \textbf{60.63} & 60.38 & \textbf{60.60} & 59.83 & 55.16 & \textbf{62.62} & - & 62.62 \\ \hdashline\noalign{\vskip 2pt}
Q3-1.7B & 61.99 & \textbf{62.36} & 62.33 & \textbf{62.44} & 59.41 & \textbf{66.31} & 59.50 & \textbf{66.59} \\
Q3-8B & \textbf{62.62} & 61.82 & \textbf{62.61} & 60.00 & 52.35 & \textbf{65.70} & - & 65.70 \\
Q3-14B & \textbf{62.50} & 61.53 & \textbf{62.44} & 59.45 & 50.82 & \textbf{65.16} & - & 65.16 \\
Q3-32B & \textbf{62.47} & 62.26 & \textbf{62.60} & 61.53 & 52.04 & \textbf{66.46} & 60.44 & \textbf{66.53} \\ \hdashline\noalign{\vskip 2pt}
Q3-VL-8B    & \textbf{61.92} & 60.10 & \textbf{61.53} & 57.20 & 47.40 & \textbf{63.54} & 60.24 & \textbf{63.61} \\
Q3-VL-32B   & \textbf{61.92} & 61.19 & \textbf{62.44} & 58.24 & 48.65 & \textbf{63.61} & 54.47 & \textbf{63.72} \\ \hdashline\noalign{\vskip 2pt}
R1-Q-14B & \textbf{62.57} & 62.27 & 62.26 & \textbf{62.28} & 56.97 & \textbf{66.29} & 64.45 & \textbf{66.37} \\
R1-Q-32B & 61.90 & \textbf{62.50} & \textbf{62.86} & 61.71 & 52.83 & \textbf{66.45} & - & 66.45 \\ \hdashline\noalign{\vskip 2pt}
R1-L-8B & 59.94 & \textbf{62.81} & 62.79 & \textbf{62.87} & 57.44 & \textbf{66.49} & 64.16 & \textbf{66.85} \\
R1-L-70B & 62.75 & \textbf{62.93} & 62.86 & \textbf{63.07} & 0.00 & \textbf{66.33} & - & 66.33 \\
\bottomrule
\end{tabular}
\caption{BERTscore $\uparrow$}
\end{subtable}
\caption{Text similarity scores (\%) between first round generations that pass or fail the quality checks.}
\label{quality_check_pass_and_not_pass}
\end{table}}

Table \ref{quality_check_fail_this_pass_next} shows text similarity scores between first round generations that fail the quality checks and their corresponding second round generations that pass them. Only second round generations that pass the \emph{Specification} check show higher ROUGE-L and BERTScore to human-written review than their first round failing counterparts.

{\setlength{\tabcolsep}{3pt}
\begin{table}[!t]
\scriptsize
\begin{subtable}{\linewidth}
\centering
\begin{tabular}{lcccccccc}
\toprule
\multirow{2.4}{*}{\textbf{model}} & \multicolumn{2}{c}{\textbf{Relevance}} & \multicolumn{2}{c}{\textbf{Specification}} & \multicolumn{2}{c}{\textbf{Evidence}} & \multicolumn{2}{c}{\textbf{Reasoning}} \\ \cmidrule(lr){2-3} \cmidrule(lr){4-5} \cmidrule(lr){6-7} \cmidrule(lr){8-9}
& \textbf{1-F} & \textbf{2-T} & \textbf{1-F} & \textbf{2-T} & \textbf{1-F} & \textbf{2-T} & \textbf{1-F} & \textbf{2-T} \\ \midrule
G-O-20B  & \textbf{19.75} & 11.08 & \textbf{17.69} & 16.69 & \textbf{18.00} & 6.55 & \textbf{42.86} & 17.04 \\
G-O-120B & \textbf{20.78} & 9.82 & \textbf{18.73} & 16.15 & \textbf{19.30} & 4.04 & - & - \\ \hdashline\noalign{\vskip 2pt}
Q3-1.7B & \textbf{27.23} & 7.96 & \textbf{30.38} & 19.54 & \textbf{26.32} & 7.10 & \textbf{25.00} & 4.96 \\
Q3-8B & \textbf{23.28} & 19.45 & \textbf{21.30} & 20.04 & \textbf{16.33} & 3.29 & - & - \\
Q3-14B & \textbf{22.07} & 15.44 & 21.10 & \textbf{21.28} & \textbf{17.83} & 4.09 & - & - \\
Q3-32B & \textbf{21.24} & 19.43 & \textbf{21.24} & 19.43 & \textbf{17.14} & 10.65 & \textbf{50.00} & 42.45 \\ \hdashline\noalign{\vskip 2pt}
Q3-VL-8B & \textbf{22.46} & 6.26 & \textbf{20.58} & 18.75 & \textbf{17.65} & 4.11 & \textbf{17.14} & 6.02 \\
Q3-VL-32B & \textbf{18.09} & 13.94 & \textbf{17.15} & 15.62 & \textbf{16.19} & 10.36 & \textbf{6.58} & 4.68 \\ \hdashline\noalign{\vskip 2pt}
R1-Q-14B & \textbf{25.23} & 12.57 & \textbf{22.47} & 17.39 & \textbf{9.10} & 8.95 & 18.87 & \textbf{21.85} \\
R1-Q-32B & \textbf{23.87} & 17.66 & \textbf{22.12} & 18.17 & \textbf{18.83} & 11.03 & - & - \\ \hdashline\noalign{\vskip 2pt}
R1-L-8B & \textbf{26.92} & 10.96 & \textbf{30.90} & 15.67 & \textbf{20.00} & 8.00 & \textbf{37.21} & 8.66 \\
R1-L-70B & \textbf{24.63} & 17.87 & \textbf{23.66} & 19.07 & \textbf{21.45} & 4.41 & - & - \\
\bottomrule
\end{tabular}
\caption{BLEU-1}
\end{subtable}

\vspace{1em}

\begin{subtable}{\linewidth}
\centering
\begin{tabular}{lcccccccc}
\toprule
\multirow{2.4}{*}{\textbf{model}} & \multicolumn{2}{c}{\textbf{Relevance}} & \multicolumn{2}{c}{\textbf{Specification}} & \multicolumn{2}{c}{\textbf{Evidence}} & \multicolumn{2}{c}{\textbf{Reasoning}} \\ \cmidrule(lr){2-3} \cmidrule(lr){4-5} \cmidrule(lr){6-7} \cmidrule(lr){8-9}
& \textbf{1-F} & \textbf{2-T} & \textbf{1-F} & \textbf{2-T} & \textbf{1-F} & \textbf{2-T} & \textbf{1-F} & \textbf{2-T} \\ \midrule
G-O-20B  & \textbf{16.61} & 12.40 & \textbf{13.73} & 14.85 & \textbf{11.54} & 8.26 & \textbf{30.56} & 19.23 \\
G-O-120B & \textbf{14.52} & 11.03 & \textbf{14.45} & 15.11 & \textbf{12.78} & 5.65 & - & - \\ \hdashline\noalign{\vskip 2pt}
Q3-1.7B & \textbf{15.38} & 9.90 & 15.12 & \textbf{15.93} & \textbf{15.49} & 9.10 & \textbf{15.09} & 8.20 \\
Q3-8B & \textbf{17.32} & 16.30 & 16.03 & \textbf{17.26} & \textbf{14.44} & 5.24 & - & - \\
Q3-14B & \textbf{15.44} & 13.97 & 15.91 & \textbf{17.10} & \textbf{12.30} & 6.07 & - & - \\
Q3-32B & \textbf{16.02} & 13.74 & 16.43 & \textbf{16.78} & \textbf{13.77} & 12.93 & 15.60 & \textbf{20.85} \\ \hdashline\noalign{\vskip 2pt}
Q3-VL-8B & \textbf{15.80} & 8.73 & 15.08 & \textbf{15.89} & \textbf{14.50} & 7.73 & \textbf{15.38} & 8.22 \\
Q3-VL-32B & \textbf{13.68} & 11.98 & 14.38 & \textbf{14.50} & \textbf{13.14} & 10.88 & \textbf{6.90} & 5.81 \\ \hdashline\noalign{\vskip 2pt}
R1-Q-14B & \textbf{18.77} & 14.05 & \textbf{16.31} & 15.32 & \textbf{16.43} & 10.62 & \textbf{14.00} & 10.56 \\
R1-Q-32B & \textbf{15.41} & 14.58 & 14.55 & \textbf{15.22} & \textbf{11.66} & 11.64 & - & - \\ \hdashline\noalign{\vskip 2pt}
R1-L-8B & \textbf{16.69} & 12.88 & \textbf{19.49} & 14.11 & \textbf{14.58} & 6.98 & \textbf{20.06} & 10.29 \\
R1-L-70B & \textbf{15.65} & 14.18 & \textbf{16.74} & 16.33 & \textbf{14.21} & 6.70 & - & - \\
\bottomrule
\end{tabular}
\caption{ROUGE-L}
\end{subtable}

\vspace{1em}

\begin{subtable}{\linewidth}
\centering
\begin{tabular}{lcccccccc}
\toprule
\multirow{2.4}{*}{\textbf{model}} & \multicolumn{2}{c}{\textbf{Relevance}} & \multicolumn{2}{c}{\textbf{Specification}} & \multicolumn{2}{c}{\textbf{Evidence}} & \multicolumn{2}{c}{\textbf{Reasoning}} \\ \cmidrule(lr){2-3} \cmidrule(lr){4-5} \cmidrule(lr){6-7} \cmidrule(lr){8-9}
& \textbf{1-F} & \textbf{2-T} & \textbf{1-F} & \textbf{2-T} & \textbf{1-F} & \textbf{2-T} & \textbf{1-F} & \textbf{2-T} \\ \midrule
G-O-20B  & \textbf{61.55} & 59.90 & 60.13 & \textbf{61.65} & \textbf{57.84} & 54.42 & \textbf{68.58} & 65.01 \\
G-O-120B & \textbf{60.62} & 59.00 & 60.55 & \textbf{61.77} & \textbf{57.47} & 53.25 & - & - \\ \hdashline\noalign{\vskip 2pt}
Q3-1.7B & \textbf{62.55} & 56.57 & \textbf{61.86} & 61.41 & \textbf{60.03} & 53.74 & \textbf{59.50} & 49.28 \\
Q3-8B & 63.57 & \textbf{64.21} & 62.26 & \textbf{63.91} & \textbf{59.18} & 54.31 & - & - \\
Q3-14B & \textbf{62.13} & 61.57 & 62.46 & \textbf{64.04} & 49.81 & \textbf{53.12} & - & - \\
Q3-32B & \textbf{61.73} & 61.33 & 62.68 & \textbf{63.53} & 54.49 & \textbf{60.16} & 60.44 & \textbf{65.10} \\ \hdashline\noalign{\vskip 2pt}
Q3-VL-8B & \textbf{61.90} & 56.21 & 61.37 & \textbf{63.01} & 51.51 & \textbf{54.31} & \textbf{60.24} & 55.48 \\
Q3-VL-32B & \textbf{61.94} & 60.92 & 62.10 & \textbf{62.84} & 55.07 & \textbf{58.24} & \textbf{54.47} & 51.68 \\ \hdashline\noalign{\vskip 2pt}
R1-Q-14B & \textbf{64.06} & 62.62 & 62.22 & \textbf{63.02} & \textbf{63.29} & 58.01 & \textbf{64.45} & 62.07 \\
R1-Q-32B & 63.02 & \textbf{63.09} & 62.82 & \textbf{63.85} & 52.51 & \textbf{59.21} & - & - \\ \hdashline\noalign{\vskip 2pt}
R1-L-8B & \textbf{62.72} & 61.59 & \textbf{63.52} & 62.45 & \textbf{59.68} & 55.44 & \textbf{64.16} & 61.74 \\
R1-L-70B & \textbf{62.47} & 62.23 & 62.79 & \textbf{63.58} & \textbf{58.86} & 55.14 & - & - \\
\bottomrule
\end{tabular}
\caption{BERTScore}
\end{subtable}
\caption{Text similarity scores (\%) between first round generations that fail the quality checks (\textbf{1-F}) and their corresponding second round generations that pass them (\textbf{2-T}).}
\label{quality_check_fail_this_pass_next}
\end{table}}

Table \ref{evaluation_scores_score_prediction_first_and_last_round} compares score prediction results based on the first round generations with those based on PoF generations.

\begin{table*}[!ht]
\scriptsize
\begin{subtable}{0.49\linewidth}
\centering
\begin{tabular}{lcccc}
\toprule
\multirow{2.4}{*}{\textbf{model}} & \multicolumn{2}{c}{\textbf{$\leq$0.5} $\uparrow$} & \multicolumn{2}{c}{\textbf{MAE} $\downarrow$} \\ \cmidrule(lr){2-3} \cmidrule(lr){4-5}
& \textbf{first} & \textbf{PoF} & \textbf{first} & \textbf{PoF} \\ \midrule
GPT-OSS-20B & \textbf{74.47} & 67.02 & \textbf{0.4894} & 0.5851 \\
GPT-OSS-120B & \textbf{76.77} & 72.72 & \textbf{0.5152} & 0.5455 \\ \hdashline\noalign{\vskip 2pt}
Qwen3-1.7B & \textbf{64.40} & 55.93 & \textbf{0.6356} & 0.7288 \\
Qwen3-8B & \textbf{53.57} & 51.19 & \textbf{0.6964} & 0.7440 \\
Qwen3-14B & \textbf{76.19} & 71.43 & \textbf{0.5119} & 0.5655 \\
Qwen3-32B & 65.22 & \textbf{68.47} & 0.5924 & \textbf{0.5707} \\ \hdashline\noalign{\vskip 2pt}
Qwen3-VL-8B & \textbf{73.33} & 70.67 & \textbf{0.5400} & 0.5733 \\
Qwen3-VL-32B & 70.00 & \textbf{75.00} & 0.5400 & \textbf{0.4950} \\ \hdashline\noalign{\vskip 2pt}
R1-Qwen-14B & \textbf{72.22} & 69.01 & 0.5556 & \textbf{0.5493} \\
R1-Qwen-32B & \textbf{65.17} & 64.13 & \textbf{0.6067} & 0.6250 \\ \hdashline\noalign{\vskip 2pt}
R1-LLaMA-8B & 41.47 & \textbf{46.34} & 0.9512 & \textbf{0.8902} \\
R1-LLaMA-70B & \textbf{69.32} & 65.52 & \textbf{0.5114} & 0.5862 \\ \midrule
\textsc{average} & \textbf{67.63} & 65.11 & \textbf{0.5854} & 0.6134 \\
\bottomrule
\end{tabular}
\caption{Soundness}
\end{subtable}
\hfill
\begin{subtable}{0.49\linewidth}
\centering
\begin{tabular}{lcccc}
\toprule
\multirow{2.4}{*}{\textbf{model}} & \multicolumn{2}{c}{\textbf{$\leq$0.5} $\uparrow$} & \multicolumn{2}{c}{\textbf{MAE} $\downarrow$} \\ \cmidrule(lr){2-3} \cmidrule(lr){4-5}
& \textbf{first} & \textbf{PoF} & \textbf{first} & \textbf{PoF} \\ \midrule
GPT-OSS-20B & \textbf{69.15} & 63.83 & \textbf{0.5691} & 0.6117 \\
GPT-OSS-120B & \textbf{76.76} & 64.64 & \textbf{0.4798} & 0.5758 \\ \hdashline\noalign{\vskip 2pt}
Qwen3-1.7B & \textbf{64.40} & 59.32 & \textbf{0.5763} & 0.7034 \\
Qwen3-8B & \textbf{55.95} & 50.00 & \textbf{0.6845} & 0.7262 \\
Qwen3-14B & \textbf{78.57} & 69.05 & \textbf{0.5179} & 0.5714 \\
Qwen3-32B & 51.09 & \textbf{56.52} & 0.6902 & \textbf{0.6467} \\ \hdashline\noalign{\vskip 2pt}
Qwen3-VL-8B & \textbf{70.66} & 66.66 & \textbf{0.5800} & 0.6067 \\
Qwen3-VL-32B & 58.00 & \textbf{61.00} & 0.6450 & \textbf{0.5950} \\ \hdashline\noalign{\vskip 2pt}
R1-Qwen-14B & \textbf{61.11} & 60.57 & 0.7431 & \textbf{0.7113} \\
R1-Qwen-32B & \textbf{70.79} & 58.70 & \textbf{0.5506} & 0.6576 \\ \hdashline\noalign{\vskip 2pt}
R1-LLaMA-8B & \textbf{41.47} & 34.15 & \textbf{0.9756} & 1.0244 \\
R1-LLaMA-70B & \textbf{59.09} & 56.33 & \textbf{0.6420} & 0.6437 \\ \midrule
\textsc{average} & \textbf{64.31} & 59.83 & \textbf{0.6230} & 0.6595 \\
\bottomrule
\end{tabular}
\caption{Overall assessment}
\end{subtable}
\caption{Comparison of review score predictions between those based on the first round generations and those based on the PoF generations.}
\label{evaluation_scores_score_prediction_first_and_last_round}
\end{table*}



Table \ref{evaluation_scores_llm_judge_dimension_first_round} shows the win rates for end-to-end and first round generations. Table \ref{evaluation_scores_llm_judge_first_last} compares the win rates of first round and PoF generations. \backtomain{more_results}

\begin{table*}[!ht]
\scriptsize
\begin{subtable}{0.49\linewidth}
\centering
\setlength{\tabcolsep}{5pt}
\begin{tabular}{lcccccc}
\toprule
\multirow{2.5}{*}{\textbf{model}} & \multicolumn{2}{c}{\textbf{TA} $\uparrow$} & \multicolumn{2}{c}{\textbf{CV} $\uparrow$} & \multicolumn{2}{c}{\textbf{AD} $\uparrow$} \\ \cmidrule(lr{3pt}){2-3}\cmidrule(lr{3pt}){4-5}\cmidrule(lr{3pt}){6-7}
& \textbf{e2e} & \textbf{jge} & \textbf{e2e} & \textbf{jge} & \textbf{e2e} & \textbf{jge} \\ \midrule
G-O-20B & \textbf{76.54} & 16.67 & \textbf{83.95} & 12.97 & \textbf{80.87} & 14.20 \\
G-O-120B & \textbf{75.53} & 15.96 & \textbf{85.64} & 11.71 & \textbf{84.57} & 12.24 \\ \hdashline\noalign{\vskip 2pt}
Q3-1.7B & \textbf{61.02} & 32.20 & \textbf{71.19} & 25.43 & \textbf{64.41} & 32.21 \\
Q3-8B & 47.03 & \textbf{48.81} & \textbf{54.76} & 43.46 & 46.43 & \textbf{51.79} \\
Q3-14B & 40.71 & \textbf{49.10} & \textbf{57.51} & 37.70 & 44.39 & \textbf{50.83} \\
Q3-32B & \textbf{52.43} & 37.72 & \textbf{65.57} & 28.42 & \textbf{59.58} & 34.95 \\ \hdashline\noalign{\vskip 2pt}
Q3-V-8B & \textbf{58.00} & 32.00 & \textbf{65.33} & 26.67 & \textbf{58.00} & 34.67 \\
Q3-V-32B & 42.16 & \textbf{48.80} & \textbf{61.80} & 33.17 & \textbf{50.25} & 44.22 \\ \hdashline\noalign{\vskip 2pt}
R1-Q-14B & 40.28 & \textbf{56.95} & 42.36 & \textbf{56.95} & 41.67 & \textbf{57.64} \\
R1-Q-32B & 27.53 & \textbf{67.98} & 33.15 & \textbf{65.17} & 25.84 & \textbf{71.91} \\ \hdashline\noalign{\vskip 2pt}
R1-L-8B & \textbf{67.50} & 30.00 & \textbf{73.75} & 25.00 & \textbf{65.00} & 32.50 \\
R1-L-70B & 21.27 & \textbf{74.14} & 25.87 & \textbf{72.99} & 19.54 & \textbf{78.74} \\ \midrule
\textsc{average} & \textbf{50.81} & 41.68 & \textbf{60.18} & 35.86 & \textbf{53.97} & 41.65 \\
\bottomrule
\end{tabular}
\caption{}
\label{evaluation_scores_llm_judge_dimension_first_round}
\end{subtable}
\hfill
\begin{subtable}{0.49\linewidth}
\centering
\begin{tabular}{lcccccc}
\toprule
\multirow{2.4}{*}{\textbf{model}} & \multicolumn{2}{c}{\textbf{TA}} & \multicolumn{2}{c}{\textbf{CV}} & \multicolumn{2}{c}{\textbf{AD}} \\ 
\cmidrule(lr){2-3}\cmidrule(lr){4-5}\cmidrule(lr){6-7}
& \textbf{first} & \textbf{PoF} & \textbf{first} & \textbf{PoF} & \textbf{first} & \textbf{PoF} \\
\midrule
G-O-20B & 16.67 & \textbf{25.93} & 12.97 & \textbf{20.99} & 14.20 & \textbf{24.69} \\
G-O-120B & 15.96 & \textbf{36.30} & 11.71 & \textbf{30.32} & 12.24 & \textbf{37.76} \\ \hdashline\noalign{\vskip 2pt}
Q3-1.7B & 32.20 & \textbf{48.31} & 25.43 & \textbf{41.53} & 32.21 & \textbf{54.24} \\
Q3-8B & \textbf{48.81} & 45.84 & \textbf{43.46} & 41.08 & 51.79 & \textbf{52.98} \\
Q3-14B & \textbf{49.10} & 46.15 & \textbf{37.70} & 36.50 & \textbf{50.83} & 47.86 \\
Q3-32B & 37.72 & \textbf{42.39} & 28.42 & \textbf{34.24} & 34.95 & \textbf{45.11} \\ \hdashline\noalign{\vskip 2pt}
Q3-VL-8B & 32.00 & \textbf{50.00} & 26.67 & \textbf{32.44} & 34.67 & \textbf{45.95} \\
Q3-VL-32B & 48.80 & \textbf{51.66} & 33.17 & \textbf{44.52} & 44.22 & \textbf{52.56} \\ \hdashline\noalign{\vskip 2pt}
R1-Q-14B & 56.95 & \textbf{66.67} & 56.95 & \textbf{65.28} & 57.64 & \textbf{71.53} \\
R1-Q-32B & 67.98 & \textbf{77.53} & 65.17 & \textbf{71.91} & 71.91 & \textbf{76.97} \\ \hdashline\noalign{\vskip 2pt}
R1-L-8B & 30.00 & \textbf{44.30} & 25.00 & \textbf{36.80} & 32.50 & \textbf{46.93} \\
R1-L-70B & 74.14 & \textbf{75.86} & 72.99 & \textbf{74.71} & 78.74 & \textbf{82.19} \\ \midrule
\textsc{average} & 41.68 & \textbf{51.41} & 35.85 & \textbf{45.63} & 41.64 & \textbf{54.01} \\
\bottomrule
\end{tabular}
\caption{}
\label{evaluation_scores_llm_judge_first_last}
\end{subtable}
\caption{(a) LLM-as-a-judge win rates (\%) for end-to-end and first round generations. Win rate is the proportion of cases where one generation is judged to be better than the other. TA = Technical Accuracy, CV = Constructive Value, AD = Analytical Depth. (b) LLM-as-a-judge win rates (\%) comparing first round generations and PoF generations against end-to-end generations.}
\end{table*}

\begin{table*}[!ht]
\small
\centering
\begin{tabular}{p{0.46\textwidth}p{0.46\textwidth}}
\toprule
\textbf{Qwen3-1.7B} & \textbf{Qwen3-32B} \\ \midrule
The methodology lacks detailed description of the data collection process and the evaluation framework, making it difficult to assess the robustness of the results. &
The task formulation for MCPQA is somewhat abstract and could benefit from more concrete examples or clearer definitions of what constitutes a ``cross-market’’ scenario. Additionally, the paper could have provided more justification for selecting specific marketplaces and languages, which might help clarify the practical relevance and coverage of the proposed dataset. \\ \midrule
The Data/Task aspect is not well-defined, and the task’s definition is somewhat ambiguous, which may affect the reproducibility of the research. &
The methodology for leveraging cross-market information is not deeply explored. For example, the paper could have discussed more nuanced approaches to aligning or translating data between languages, or how the model handles variations in product metadata and user-generated content across different marketplaces. \\ \midrule
The evaluation is not comprehensive, with limited analysis of the performance metrics and their interpretation in the context of the task. &
The evaluation section, while extensive, lacks some depth in analyzing the limitations of the proposed methods. For instance, the paper could have compared its results with more established cross-lingual QA baselines or explored the impact of translation errors on performance. Additionally, the paper does not discuss how the proposed methods might generalize to other languages or marketplaces not included in the dataset. \\
\bottomrule
\end{tabular}
\caption{Example judgment-grounded expansion outputs from Qwen3-1.7B and Qwen3-32B for the same input.}
\label{case_study_underperforming_jge}
\end{table*}

\subsubsection{Manual analysis}
\label{manual_analysis}

We inspect cases where judgment-grounded expansion underperforms end-to-end generation. We focus on Qwen3-1.7B, which shows the largest gap. We find that this model expands reviewer judgment in a generic way without grounding the comment in the submission. As shown in Table \ref{case_study_underperforming_jge}, Qwen3-1.7B mostly produces broad comments and does not connect them to concrete paper details. In contrast, Qwen3-32B produces more paper-specific expansions that refer to specific elements of the submission. This suggests that the gains from reviewer judgment depend on whether the model can effectively use the provided judgment. It highlights the need for large-scale evaluation of models' \emph{judgment-utilization ability}. \backtomain{more_results}

\setreturn{other_results}
\subsubsection{Other results}
\label{other_results}

Tables \ref{evaluation_scores_similarity_iclr25}, \ref{evaluation_scores_score_prediction_iclr25}, and \ref{evaluation_scores_llm_judge_dimension_iclr25} show ICLR results.

\begin{table*}[!ht]
\scriptsize
{\settowidth{\DeltaColW}{00.00}
\DeltaSetRange{-25}{45}
\begin{subtable}{0.3\linewidth}
\centering
\begin{tabular}{lccc}
\toprule
\textbf{model} & \textbf{e2e} & \textbf{jge} & \textbf{$\Delta$\%} \\ \midrule
G-O-20B   & 0.0223 & \textbf{0.0247} & \DeltaHeat{+10.76} \\
G-O-120B  & 0.0241 & \textbf{0.0286} & \DeltaHeat{+18.67} \\ \hdashline\noalign{\vskip 2pt}
Q3-14B    & \textbf{0.0401} & 0.0397 & \DeltaHeat{-1.00} \\
Q3-32B    & \textbf{0.0390} & 0.0369 & \DeltaHeat{-5.38} \\ \hdashline\noalign{\vskip 2pt}
R1-Q-14B  & 0.0389 & \textbf{0.0402} & \DeltaHeat{+3.34} \\
R1-Q-32B  & \textbf{0.0425} & 0.0374 & \DeltaHeat{-12.00} \\
\bottomrule
\end{tabular}
\caption{BLEU $\uparrow$}
\end{subtable}
\hfill
\DeltaSetRange{-1}{16}
\begin{subtable}{0.3\linewidth}
\centering
\begin{tabular}{lccc}
\toprule
\textbf{model} & \textbf{e2e} & \textbf{jge} & \textbf{$\Delta$\%} \\ \midrule
G-O-20B   & 0.1464 & \textbf{0.1582} & \DeltaHeat{+8.06} \\
G-O-120B  & 0.1485 & \textbf{0.1672} & \DeltaHeat{+12.59} \\ \hdashline\noalign{\vskip 2pt}
Q3-14B    & 0.1792 & \textbf{0.1813} & \DeltaHeat{+1.17} \\
Q3-32B    & 0.1729 & \textbf{0.1756} & \DeltaHeat{+1.56} \\ \hdashline\noalign{\vskip 2pt}
R1-Q-14B  & 0.1684 & \textbf{0.1806} & \DeltaHeat{+7.24} \\
R1-Q-32B  & 0.1724 & \textbf{0.1768} & \DeltaHeat{+2.55} \\
\bottomrule
\end{tabular}
\caption{ROUGE-L $\uparrow$}
\end{subtable}
\hfill
\DeltaSetRange{-0.5}{3.5}
\begin{subtable}{0.3\linewidth}
\centering
\begin{tabular}{lccc}
\toprule
\textbf{model} & \textbf{e2e} & \textbf{jge} & \textbf{$\Delta$\%} \\ \midrule
G-O-20B   & 0.6444 & \textbf{0.6534} & \DeltaHeat{+1.40} \\
G-O-120B  & 0.6463 & \textbf{0.6573} & \DeltaHeat{+1.70} \\ \hdashline\noalign{\vskip 2pt}
Q3-14B    & \textbf{0.6800} & 0.6776 & \DeltaHeat{-0.35} \\
Q3-32B    & \textbf{0.6761} & 0.6756 & \DeltaHeat{-0.07} \\ \hdashline\noalign{\vskip 2pt}
R1-Q-14B  & 0.6702 & \textbf{0.6798} & \DeltaHeat{+1.43} \\
R1-Q-32B  & 0.6738 & \textbf{0.6778} & \DeltaHeat{+0.59} \\
\bottomrule
\end{tabular}
\caption{BERTScore $\uparrow$}
\end{subtable}}
\caption{Text similarity scores between end-to-end generations (\textbf{e2e}), judgment-grounded expansions (\textbf{jge}), and human-written reviews for the ICLR data. G-O = GPT-OSS, Q = Qwen, L = LLaMA, V = VL, DeepR = DeepReview. \backtoappendix{other_results}}
\label{evaluation_scores_similarity_iclr25}
\end{table*}

\begin{table*}[!ht]
\scriptsize
\setlength{\tabcolsep}{1pt}

\begin{subtable}{0.24\linewidth}
\centering
\begin{tabular}{lcccc}
\toprule
\multirow{2.4}{*}{\textbf{model}} & \multicolumn{2}{c}{\textbf{EM} $\uparrow$} & \multicolumn{2}{c}{\textbf{MAE} $\downarrow$} \\
\cmidrule(lr){2-3} \cmidrule(lr){4-5}
& \textbf{e2e} & \textbf{jge} & \textbf{e2e} & \textbf{jge} \\
\midrule
G-O-20B  & \textbf{66.23} & 64.21 & \textbf{0.3377} & 0.3895 \\
G-O-120B & 53.00 & \textbf{55.56} & 0.4800 & \textbf{0.4747} \\
\hdashline\noalign{\vskip 2pt}
Q3-14B    & \textbf{64.00} & 61.29 & \textbf{0.3800} & 0.4086 \\
Q3-32B    & 44.00 & \textbf{57.14} & 0.6900 & \textbf{0.5000} \\
\hdashline\noalign{\vskip 2pt}
R1-Q-14B  & \textbf{63.92} & 62.34 & \textbf{0.3918} & 0.4026 \\
R1-Q-32B  & 55.43 & \textbf{62.92} & 0.5000 & \textbf{0.4157} \\
\bottomrule
\end{tabular}
\caption{Soundness}
\end{subtable}
\hfill
\begin{subtable}{0.24\linewidth}
\centering
\begin{tabular}{lcccc}
\toprule
\multirow{2.4}{*}{\textbf{model}} & \multicolumn{2}{c}{\textbf{EM} $\uparrow$} & \multicolumn{2}{c}{\textbf{MAE} $\downarrow$} \\
\cmidrule(lr){2-3} \cmidrule(lr){4-5}
& \textbf{e2e} & \textbf{jge} & \textbf{e2e} & \textbf{jge} \\
\midrule
G-O-20B  & \textbf{51.95} & 51.58 & \textbf{0.5195} & 0.6105 \\
G-O-120B & 45.00 & \textbf{51.52} & 0.6400 & \textbf{0.5455} \\
\hdashline\noalign{\vskip 2pt}
Q3-14B    & 69.00 & \textbf{69.89} & 0.3400 & \textbf{0.3226} \\
Q3-32B    & \textbf{62.00} & 60.20 & \textbf{0.4100} & 0.4286 \\
\hdashline\noalign{\vskip 2pt}
R1-Q-14B  & \textbf{68.04} & 64.94 & \textbf{0.3711} & 0.4026 \\
R1-Q-32B  & 48.91 & \textbf{61.80} & 0.5870 & \textbf{0.4494} \\
\bottomrule
\end{tabular}
\caption{Presentation}
\end{subtable}
\hfill
\begin{subtable}{0.24\linewidth}
\centering
\begin{tabular}{lcccc}
\toprule
\multirow{2.4}{*}{\textbf{model}} & \multicolumn{2}{c}{\textbf{EM} $\uparrow$} & \multicolumn{2}{c}{\textbf{MAE} $\downarrow$} \\
\cmidrule(lr){2-3} \cmidrule(lr){4-5}
& \textbf{e2e} & \textbf{jge} & \textbf{e2e} & \textbf{jge} \\
\midrule
G-O-20B  & 28.57 & \textbf{32.63} & 0.8571 & \textbf{0.7368} \\
G-O-120B & 46.00 & \textbf{48.48} & 0.5900 & \textbf{0.5354} \\
\hdashline\noalign{\vskip 2pt}
Q3-14B    & 43.00 & \textbf{43.01} & 0.6000 & \textbf{0.5806} \\
Q3-32B    & 31.00 & \textbf{39.80} & 0.8200 & \textbf{0.6327} \\
\hdashline\noalign{\vskip 2pt}
R1-Q-14B  & \textbf{38.14} & 37.66 & 0.7216 & \textbf{0.7013} \\
R1-Q-32B  & 34.78 & \textbf{41.57} & 0.7391 & \textbf{0.6404} \\
\bottomrule
\end{tabular}
\caption{Contribution}
\end{subtable}
\hfill
\begin{subtable}{0.24\linewidth}
\centering
\begin{tabular}{lcccc}
\toprule
\multirow{2.4}{*}{\textbf{model}} & \multicolumn{2}{c}{\textbf{EM} $\uparrow$} & \multicolumn{2}{c}{\textbf{MAE} $\downarrow$} \\
\cmidrule(lr){2-3} \cmidrule(lr){4-5}
& \textbf{e2e} & \textbf{jge} & \textbf{e2e} & \textbf{jge} \\
\midrule
G-O-20B  & 15.58 & \textbf{30.53} & 2.1169 & \textbf{1.6211} \\
G-O-120B & 22.00 & \textbf{25.25} & 1.7700 & \textbf{1.6364} \\
\hdashline\noalign{\vskip 2pt}
Q3-14B    & 8.00 & \textbf{10.75} & 2.6200 & \textbf{2.4839} \\
Q3-32B    & 8.00 & \textbf{9.18} & 2.9800 & \textbf{2.5408} \\
\hdashline\noalign{\vskip 2pt}
R1-Q-14B  & 13.40 & \textbf{14.29} & 2.1753 & \textbf{2.0390} \\
R1-Q-32B  & 8.70 & \textbf{14.61} & 2.7283 & \textbf{2.4270} \\
\bottomrule
\end{tabular}
\caption{Rating}
\end{subtable}

\caption{Exact match (\textbf{EM}) and mean absolute error (\textbf{MAE}) between score predictions and the corresponding human-assigned scores for the ICLR data. \backtoappendix{other_results}}
\label{evaluation_scores_score_prediction_iclr25}
\end{table*}

{\setlength{\tabcolsep}{4pt}
\begin{table*}[!t]
\scriptsize
\centering
\begin{tabular}{lcccccc}
\toprule
\multirow{2.4}{*}{\textbf{model}} & \multicolumn{2}{c}{\textbf{TA}} & \multicolumn{2}{c}{\textbf{CV}} & \multicolumn{2}{c}{\textbf{AD}} \\ 
\cmidrule(lr){2-3}\cmidrule(lr){4-5}\cmidrule(lr){6-7}
& \textbf{e2e} & \textbf{jge} & \textbf{e2e} & \textbf{jge} & \textbf{e2e} & \textbf{jge} \\ 
\midrule
G-O-20B   & \textbf{73.65} & 17.57 & \textbf{79.06} & 18.25 & \textbf{77.03} & 20.27 \\
G-O-120B  & \textbf{56.97} & 34.82 & \textbf{62.70} & 33.72 & \textbf{59.14} & 36.77 \\ \hdashline\noalign{\vskip 2pt}
Q3-14B     & 34.95 & \textbf{55.91} & 43.02 & \textbf{52.15} & 36.56 & \textbf{59.14} \\
Q3-32B     & \textbf{48.47} & 39.80 & \textbf{54.59} & 37.76 & \textbf{46.94} & 46.43 \\ \hdashline\noalign{\vskip 2pt}
R1-Q-14B   & 31.58 & \textbf{65.79} & 40.79 & \textbf{57.90} & 34.21 & \textbf{64.48} \\
R1-Q-32B   & 25.89 & \textbf{71.77} & 28.82 & \textbf{68.83} & 23.53 & \textbf{74.12} \\ \midrule
\textsc{average} & 45.25 & \textbf{47.61} & \textbf{51.49} & 44.76 & 46.23 & \textbf{50.20} \\
\bottomrule
\end{tabular}
\caption{LLM-as-a-judge win rates (\%) for end-to-end and judgment-grounded expansion across the three LLM judge criteria for the ICLR data. Win rate is the proportion of cases where one generation is judged to be better than the other. TA = Technical Accuracy, CV = Constructive Value, AD = Analytical Depth. \backtoappendix{more_on_recommendation_accuracy}}
\label{evaluation_scores_llm_judge_dimension_iclr25}
\end{table*}}

Table \ref{evaluation_scores_similarity_2_proxies} shows text similarity scores between judgment-grounded expansions using the exemplar-based proxy, aspect-based proxy, and human-written reviews.

\begin{table*}[!ht]
\scriptsize
{\settowidth{\DeltaColW}{00.00}
\begin{subtable}{0.3\linewidth}
\centering
\begin{tabular}{lcc}
\toprule
\textbf{model} & \textbf{exp} & \textbf{asp} \\ \midrule
G-O-20B   & \textbf{0.0341} & 0.0300 \\
G-O-120B  & 0.0270 & \textbf{0.0287} \\ \hdashline\noalign{\vskip 2pt}
Q3-14B    & \textbf{0.0668} & 0.0443 \\
Q3-32B    & \textbf{0.0602} & 0.0469 \\ \hdashline\noalign{\vskip 2pt}
R1-Q-14B  & \textbf{0.0611} & 0.0491 \\
R1-Q-32B  & \textbf{0.0533} & 0.0485 \\
\bottomrule
\end{tabular}
\caption{BLEU $\uparrow$}
\end{subtable}
\hfill
\begin{subtable}{0.3\linewidth}
\centering
\begin{tabular}{lcc}
\toprule
\textbf{model} & \textbf{exp} & \textbf{asp} \\ \midrule
G-O-20B   & \textbf{0.1702} & 0.1671 \\
G-O-120B  & \textbf{0.1726} & 0.1708 \\ \hdashline\noalign{\vskip 2pt}
Q3-14B    & \textbf{0.2277} & 0.1860 \\
Q3-32B    & \textbf{0.2213} & 0.1887 \\ \hdashline\noalign{\vskip 2pt}
R1-Q-14B  & \textbf{0.2208} & 0.1884 \\
R1-Q-32B  & \textbf{0.2048} & 0.1882 \\
\bottomrule
\end{tabular}
\caption{ROUGE-L $\uparrow$}
\end{subtable}
\hfill
\begin{subtable}{0.3\linewidth}
\centering
\begin{tabular}{lcc}
\toprule
\textbf{model} & \textbf{exp} & \textbf{asp} \\ \midrule
G-O-20B   & 0.6482 & \textbf{0.6547} \\
G-O-120B  & 0.6501 & \textbf{0.6568} \\ \hdashline\noalign{\vskip 2pt}
Q3-14B    & \textbf{0.6978} & 0.6772 \\
Q3-32B    & \textbf{0.6927} & 0.6795 \\ \hdashline\noalign{\vskip 2pt}
R1-Q-14B  & \textbf{0.6952} & 0.6785 \\
R1-Q-32B  & \textbf{0.6906} & 0.6789 \\
\bottomrule
\end{tabular}
\caption{BERTScore $\uparrow$}
\end{subtable}}
\caption{Text similarity scores between judgment-grounded expansions using the exemplar-based proxy (\textbf{exp}), aspect-based proxy (\textbf{asp}), and human-written reviews. \backtoappendix{other_results} \backtomain{sanity_check}}
\label{evaluation_scores_similarity_2_proxies}
\end{table*}

Table \ref{evaluation_scores_score_prediction_adversarial_full} shows review score prediction results under adversarial and non-adversarial settings. \backtomain{more_results}

{\setlength{\tabcolsep}{3pt}
\begin{table*}[ht]
\scriptsize
\centering
\begin{subtable}{0.48\linewidth}
\centering
\begin{tabular}{lccccccc}
\toprule
\multirow{2.4}{*}{\textbf{model}} & \multirow{2}{*}{\textbf{type}} & \multicolumn{3}{c}{\textbf{$\leq$0.5}} & \multicolumn{3}{c}{\textbf{MAE}} \\ \cmidrule(lr){3-5} \cmidrule(lr){6-8}
&  & \textbf{False} & \textbf{True} & \textbf{|$\Delta$\%| $\downarrow$} & \textbf{False} & \textbf{True} & \textbf{|$\Delta$\%| $\downarrow$} \\ \midrule
\multirow{2}{*}{G-O-20B} & \textbf{e2e} & 76.47 & 17.70 & 76.85 & 0.5353 & 1.3229 & 147.13 \\ 
& \textbf{jge} & 74.47 & 38.54 & \textbf{48.25} & 0.4894 & 1.0000 & \textbf{104.33} \\ \midrule
\multirow{2}{*}{G-O-120B} & \textbf{e2e} & 70.84 & 43.00 & 39.30 & 0.5677 & 0.9800 & 72.63 \\ 
& \textbf{jge} & 76.77 & 50.00 & \textbf{34.87} & 0.5152 & 0.7800 & \textbf{51.40} \\ \midrule
\multirow{2}{*}{Q3-1.7B} & \textbf{e2e} & 75.26 & 38.54 & 48.79 & 0.4948 & 1.0156 & 105.25 \\ 
& \textbf{jge} & 64.40 & 46.97 & \textbf{27.07} & 0.6356 & 0.8561 & \textbf{34.69} \\ \midrule
\multirow{2}{*}{Q3-8B} & \textbf{e2e} & 59.79 & 31.96 & 46.55 & 0.6701 & 1.0722 & 60.01 \\ 
& \textbf{jge} & 53.57 & 52.33 & \textbf{2.31} & 0.6964 & 0.7674 & \textbf{10.20} \\ \midrule
\multirow{2}{*}{Q3-14B} & \textbf{e2e} & 82.47 & 54.54 & 33.87 & 0.4742 & 0.7071 & 49.11 \\ 
& \textbf{jge} & 76.19 & 66.67 & \textbf{12.50} & 0.5119 & 0.5833 & \textbf{13.95} \\ \midrule
\multirow{2}{*}{Q3-32B} & \textbf{e2e} & 61.86 & 31.63 & 48.87 & 0.6186 & 1.1224 & 81.44 \\ 
& \textbf{jge} & 65.22 & 58.70 & \textbf{10.00} & 0.5924 & 0.6522 & \textbf{10.09} \\ \midrule
\multirow{2}{*}{R1-Q-14B} & \textbf{e2e} & 68.05 & 58.33 & \textbf{14.28} & 0.5619 & 0.7292 & 29.77 \\ 
& \textbf{jge} & 72.22 & 60.53 & 16.19 & 0.5556 & 0.6382 & \textbf{14.87} \\ \midrule
\multirow{2}{*}{R1-Q-32B} & \textbf{e2e} & 63.44 & 48.93 & 22.87 & 0.6022 & 0.8191 & 36.02 \\ 
& \textbf{jge} & 65.17 & 52.94 & \textbf{18.77} & 0.6067 & 0.7059 & \textbf{16.35} \\ \midrule
\multirow{2}{*}{R1-L-8B} & \textbf{e2e} & 55.55 & 43.00 & 22.59 & 0.7677 & 0.9150 & 19.19 \\ 
& \textbf{jge} & 41.47 & 32.44 & \textbf{21.77} & 0.9512 & 1.0946 & \textbf{15.08} \\ \midrule
\multirow{2}{*}{R1-L-70B} & \textbf{e2e} & 69.39 & 49.48 & 28.69 & 0.5612 & 0.8299 & 47.88 \\ 
& \textbf{jge} & 69.32 & 59.56 & \textbf{14.08} & 0.5114 & 0.6742 & \textbf{31.83} \\ 
\bottomrule
\end{tabular}
\caption{Soundness}
\end{subtable}
\hfill
\begin{subtable}{0.48\linewidth}
\centering
\begin{tabular}{lccccccc}
\toprule
\multirow{2.4}{*}{\textbf{model}} & \multirow{2}{*}{\textbf{type}} & \multicolumn{3}{c}{\textbf{$\leq$0.5}} & \multicolumn{3}{c}{\textbf{MAE}} \\ \cmidrule(lr){3-5} \cmidrule(lr){6-8}
&  & \textbf{False} & \textbf{True} & \textbf{|$\Delta$\%| $\downarrow$} & \textbf{False} & \textbf{True} & \textbf{|$\Delta$\%| $\downarrow$} \\ \midrule
\multirow{2}{*}{G-O-20B} & \textbf{e2e} & 64.71 & 8.33 & 87.13 & 0.6353 & 1.5938 & 150.87 \\ \
& \textbf{jge} & 69.15 & 30.21 & \textbf{56.31} & 0.5691 & 1.1458 & \textbf{101.34} \\ \midrule
\multirow{2}{*}{G-O-120B} & \textbf{e2e} & 66.67 & 36.00 & 46.00 & 0.6042 & 1.0100 & \textbf{67.16} \\ \
& \textbf{jge} & 76.76 & 45.00 & \textbf{41.38} & 0.4798 & 0.8200 & 70.90 \\ \midrule
\multirow{2}{*}{Q3-1.7B} & \textbf{e2e} & 64.95 & 45.83 & 29.44 & 0.5928 & 0.7656 & 29.15 \\ 
& \textbf{jge} & 64.40 & 50.00 & \textbf{22.36} & 0.5763 & 0.7348 & \textbf{27.50} \\ \midrule
\multirow{2}{*}{Q3-8B} & \textbf{e2e} & 54.64 & 27.83 & 49.07 & 0.6546 & 1.1701 & 78.75 \\ 
& \textbf{jge} & 55.95 & 47.67 & \textbf{14.80} & 0.6845 & 0.7849 & \textbf{14.67} \\ \midrule
\multirow{2}{*}{Q3-14B} & \textbf{e2e} & 81.44 & 37.37 & 54.11 & 0.4794 & 1.0404 & 117.02 \\ 
& \textbf{jge} & 78.57 & 63.34 & \textbf{19.38} & 0.5179 & 0.6000 & \textbf{15.85} \\ \midrule
\multirow{2}{*}{Q3-32B} & \textbf{e2e} & 48.45 & 15.30 & 68.42 & 0.7423 & 1.4031 & 89.02 \\ 
& \textbf{jge} & 51.09 & 43.48 & \textbf{14.90} & 0.6902 & 0.7989 & \textbf{15.75} \\ \midrule
\multirow{2}{*}{R1-Q-14B} & \textbf{e2e} & 50.52 & 42.71 & \textbf{15.46} & 0.7680 & 0.9896 & 28.85 \\ 
& \textbf{jge} & 61.11 & 50.00 & 18.18 & 0.7431 & 0.8684 & \textbf{16.86} \\ \midrule
\multirow{2}{*}{R1-Q-32B} & \textbf{e2e} & 67.74 & 35.11 & 48.17 & 0.5753 & 0.9734 & 69.20 \\ 
& \textbf{jge} & 70.79 & 51.76 & \textbf{26.88} & 0.5506 & 0.7000 & \textbf{27.13} \\ \midrule
\multirow{2}{*}{R1-L-8B} & \textbf{e2e} & 39.39 & 36.00 & \textbf{8.61} & 0.9596 & 1.0550 & \textbf{9.94} \\ 
& \textbf{jge} & 41.47 & 27.02 & 34.84 & 0.9756 & 1.2432 & 27.43 \\ \midrule
\multirow{2}{*}{R1-L-70B} & \textbf{e2e} & 51.02 & 31.96 & 37.36 & 0.6990 & 1.1289 & 61.50 \\ 
& \textbf{jge} & 59.09 & 43.82 & \textbf{25.84} & 0.6420 & 0.8090 & \textbf{26.01} \\ 
\bottomrule
\end{tabular}
\caption{Overall assessment}
\end{subtable}
\caption{Review score prediction results under adversarial and non-adversarial settings across models. Each $\Delta$ indicates the relative change. \backtoappendix{other_results}}
\label{evaluation_scores_score_prediction_adversarial_full}
\end{table*}}

\setreturn{more_on_balancing_size_and_coverage}
\subsection{More on balancing size and coverage}
\label{more_on_balancing_size_and_coverage}

We use \texttt{all-MiniLM-L6-v2} and calculate the cosine similarity between the human input and each review comment candidate \cite{reimers2019sentencebert}. We use \texttt{facebook/bart-large-mnli} to calculate the entailment score between the human input and each review comment.




\begin{table*}[!ht]
\small
\centering
\begin{tabular}{p{0.95\linewidth}}
\toprule
\texttt{Given a research paper and the review guidelines below, write a summary of its strengths and weaknesses. Output a json dictionary.}\\
\\
\texttt{\#\# Review guidelines}\\
\\
\texttt{**Summary of Strengths**}\\
\texttt{What are the major reasons to publish this paper at a selective *ACL venue? These could include novel and useful methodology, insightful empirical results or theoretical analysis, clear organization of related literature, or any other reason why interested readers of *ACL papers may find the paper useful.}\\
\\
\texttt{**Summary of Weaknesses**}\\
\texttt{What are the concerns that you have about the paper that would cause you to favor prioritizing other high-quality papers that are also under consideration for publication? These could include concerns about correctness of the results or argumentation, limited perceived impact of the methods or findings (note that impact can be significant both in broad or in narrow sub-fields), lack of clarity in exposition, or any other reason why interested readers of *ACL papers may gain less from this paper than they would from other papers under consideration. Where possible, please number your concerns so authors may respond to them individually.}\\
\\
\texttt{\#\# Output format}\\
\texttt{Output only the json dictionary and follow the json schema exactly, with no extra keys, notes, comments, or explanations:}
\texttt{\{``strengths'': ``...'', ``weaknesses'': ``...''\}} \\
\bottomrule
\end{tabular}
\caption{The prompt for end-to-end review generation. \backtoappendix{more_on_experimental_settings} \backtomain{review_generation}}
\label{end_to_end_prompt}
\end{table*}

\begin{table*}[!ht]
\small
\centering
\begin{tabular}{p{0.95\linewidth}}
\toprule
\texttt{Given a research paper and the review guidelines below, write a summary of its strengths and weaknesses. Output a json dictionary.}\\
\\
\texttt{You will also be given a dictionary of bullet points (each corresponding to a single strength or weakness), and each bullet point is associated with one or more aspects (e.g., Methodology). Your task is judgment-grounded expansion, which is to generate a comment for the paper that expands the given aspects.}\\
\\
\texttt{The number of output bullet points must match the input dictionary exactly, and each generated comment should go into the corresponding position in the output dictionary. For example, you will receive an input dictionary like this: \{``strengths'': \{``0'': [``Data/Task''], ``1'': [``Result'', ``Experiment'']\}, ``weaknesses'': \{``0'': [``Presentation''], ``1'': [``Methodology''], ``2'': [``Data/Task'', ``Result'']\}\}. This means, you must generate 2 comments for strength: the first based on Data/Task, and the second on Result and Experiment. Then, generate 3 comments for weakness: the first based on Presentation, the second on Methodology, and the third on Data/Task and Result. The final output should be:}\\
\\
\texttt{\{}\\
\texttt{\ \ \ \ ``strengths'': \{}\\
\texttt{\ \ \ \ \ \ \ \ ``0'': ``...'', \# Comment about Data/Task}\\
\texttt{\ \ \ \ \ \ \ \ ``1'': ``...''\ \ \# Comment about Result and Experiment}\\
\texttt{\ \ \ \ \ \ \ \ \},}\\
\texttt{\ \ \ \ ``weaknesses'': \{}\\
\texttt{\ \ \ \ \ \ \ \ ``0'': ``...'', \# Comment about Presentation}\\
\texttt{\ \ \ \ \ \ \ \ ``1'': ``...'', \# Comment about Methodology}\\
\texttt{\ \ \ \ \ \ \ \ ``2'': ``...''\ \ \# Comment about Data/Task and Result}\\
\texttt{\ \ \ \ \ \ \ \ \}}\\
\texttt{\}}\\
\\
\texttt{\#\# Review guidelines}\\
\\
\texttt{**Summary of Strengths**}\\
\texttt{What are the major reasons to publish this paper at a selective *ACL venue? These could include novel and useful methodology, insightful empirical results or theoretical analysis, clear organization of related literature, or any other reason why interested readers of *ACL papers may find the paper useful.}\\
\\
\texttt{**Summary of Weaknesses**}\\
\texttt{What are the concerns that you have about the paper that would cause you to favor prioritizing other high-quality papers that are also under consideration for publication? These could include concerns about correctness of the results or argumentation, limited perceived impact of the methods or findings (note that impact can be significant both in broad or in narrow sub-fields), lack of clarity in exposition, or any other reason why interested readers of *ACL papers may gain less from this paper than they would from other papers under consideration. Where possible, please number your concerns so authors may respond to them individually.}\\
\\
\texttt{\#\# Output format}\\
\texttt{Output only the json dictionary and follow the json schema exactly, with no extra keys, notes, comments, or explanations:}
\texttt{\{``strengths'': \{``0'': ``...'', ``1'': ``...'', ...\}, ``weaknesses'': \{``0'': ``...'', ``1'': ``...'', ...\}\}} \\
\bottomrule
\end{tabular}
\caption{The prompt for judgment-grounded expansion. We use the aspect-based proxy as an example. For the exemplar-based proxy, only the aspect field in the input dictionary is replaced with the exemplar-based proxy. All other parts of the prompt remain unchanged. \backtoappendix{more_on_experimental_settings} \backtomain{review_generation}}
\label{collaborative_prompt}
\end{table*}

\begin{table*}[!ht]
\small
\centering
\begin{tabular}{p{0.95\linewidth}}
\toprule
\colorbox{lightgrey}{\textbf{Relevance}}\\[1.5ex]
\texttt{This review comment does not correctly address its corresponding key point and judgment: “\$[GENERATED\_REVIEW]\$”. Re-generate the review comment from scratch. Make sure the new review comment correctly addresses its corresponding key point and judgment.}\\
\\
\texttt{\#\# Output format}\\
\texttt{Output only the re-generated review comment as a json dictionary and follow the json schema below exactly, with no extra keys, notes, or comments:}\\
\texttt{\{“review”: “...” \# the re-generated review comment\}}\\ \midrule
\colorbox{lightgrey}{\textbf{Specification}}\\[1.5ex]
\texttt{This review comment does not capture the intended specificity of its corresponding key point and judgment: “\$[GENERATED\_REVIEW]\$”. Re-generate the review comment based on the more specific key point and judgment: “\$[KEY\_POINT]\$”.}\\
\\
\texttt{\#\# Output format}\\
\texttt{Output only the re-generated review comment as a json dictionary and follow the json schema below exactly, with no extra keys, notes, or comments:}\\
\texttt{\{“review”: “...” \# the re-generated review comment\}}\\
\midrule
\colorbox{lightgrey}{\textbf{Evidence}}\\[1.5ex]
\texttt{This review comment has already provided a correct interpretation and specification of its corresponding key point and judgment, but the evidence used is problematic: “\$[GENERATED\_REVIEW]\$”. The evidence is insufficient or hallucinated.}\\
\\
\texttt{Revise the review comment by keeping the original key point and judgment unchanged, but replacing the original evidence to make it stronger and more faithful. Do not alter the interpretation and specification of its corresponding key point and judgment.}\\
\\
\texttt{\#\# Output format}\\
\texttt{Output only the revised review comment as a json dictionary and follow the json schema below exactly, with no extra keys, notes, or comments:}\\
\texttt{\{“review”: “...” \# the revised review comment\}}\\
\midrule
\colorbox{lightgrey}{\textbf{Reasoning}}\\[1.5ex]
\texttt{This review comment has already provided a correct interpretation, specification, and evidence for its corresponding key point and judgment, but the reasoning is insufficient or unclear: “\$[GENERATED\_REVIEW]\$”. The logical connection between the generated review comment and its corresponding key point and judgment is not well developed.}\\
\\
\texttt{Revise the review comment by keeping the original key point, judgment, and evidence unchanged, but elaborating the reasoning to make it clearer and more logically connected to the provided key point and judgment. Do not alter the interpretation, specification, or evidence for its corresponding key point and judgment.}\\
\\
\texttt{\#\# Output format}\\
\texttt{Output only the revised review comment as a json dictionary and follow the json schema below exactly, with no extra keys, notes, or comments:}\\
\texttt{\{“review”: “...” \# the revised review comment\}}\\
\bottomrule
\end{tabular}
\caption{The prompt for judgment-grounded expansion refinement. The full model input includes one of these prompts together with all inputs and outputs from the previous round. \backtoappendix{more_on_experimental_settings} \backtomain{review_generation}}
\label{collaborative_prompts_next_round}
\end{table*}

\begin{table*}[!ht]
\small
\centering
\begin{tabular}{p{0.95\linewidth}}
\toprule
\texttt{Simulate reviewer-provided evaluative claims for a judgment-grounded expansion setting.}\\
\\
\texttt{In this setting, a human reviewer provides a short evaluative claim about a paper which is essentially a key point plus judgment (e.g., "Novelty is a strength"), and an LLM expands it into a full review comment. Your task is the reverse: given a full review comment, generate the evaluative claim that could have plausibly led to it.}\\
\\
\texttt{The simulated evaluative claim must:}\\
\texttt{- Be short (typically one short sentence).}\\
\texttt{- Match the style, register, and granularity of the provided exemplars below.}\\
\texttt{- Not simply restate the full review comment.}\\
\\
\texttt{You will be given a batch of full review comments. For each one, output the simulated evaluative claim.}\\
\\
\texttt{\#\# Output format}\\
\texttt{Output only the json dictionary and follow the json schema exactly, with no extra keys, notes, comments, or explanations:}\\
\texttt{\{"0": "...", "1": "...", ...\}} \\
\bottomrule
\end{tabular}
\caption{The prompt for creating exemplar-based proxies. \backtoappendix{more_on_experimental_settings} \backtomain{simulating_human_input}}
\label{creating_exemplar_based_proxy}
\end{table*}

\begin{table*}[!ht]
\small
\centering
\begin{tabular}{p{0.95\linewidth}}
\toprule
\colorbox{lightgrey}{\textbf{Evidence}}\\[1.5ex]
\texttt{Evaluate whether an LLM-generated review provides appropriate and sufficient evidence to support a given key point and judgment.}\\
\\
\texttt{You will be given the following inputs:}\\
\texttt{1. Key point and judgment: This is the original input that guided the LLM's generation (e.g., ``Clarity is a weakness'').}\\
\texttt{2. LLM-generated review: The review generated by an LLM.}\\
\texttt{3. Human-written review: The original review written by a human reviewer for the same paper. This is the reference for comparison.}\\
\\
\texttt{\#\# Step 1: Determine if the review comment requires evidence.}\\
\texttt{Some review comments do not require evidence (e.g., comments describing missing or absent content like ``The paper lacks human evaluation.'' To determine whether this is the case, also inspect the human-written review. If this is the case, label it as: **NA**.}\\
\\
\texttt{\#\# Step 2: If evidence is required, check for hallucination.}\\
\texttt{If any part of the evidence is fabricated (e.g., content that does not exist in the paper, misrepresents the original content), label it as: **HALLUCINATED**.}\\
\\
\texttt{\#\# Step 3: If evidence is required and not hallucinated, perform the following analysis:}\\
\texttt{1. **MATCH** - The LLM-generated review includes evidence that matches or closely resembles the evidence in the human-written review.}\\
\texttt{2. **SUFFICIENT** - The LLM's evidence does not match the human review, but it is still relevant, specific, and adequate to support the key point and judgment.}\\
\texttt{3. **INSUFFICIENT** - The LLM's evidence neither matches the human review nor sufficiently supports the key point and judgment. It is vague, generic, or irrelevant.}\\
\\
\texttt{\#\# Output format}\\
\texttt{Output only the json dictionary and follow the json schema exactly, with no extra keys, notes, or comments:}\\
\texttt{\{``label'': ``NA'' | ``HALLUCINATED'' | ``MATCH'' | ``SUFFICIENT'' | ``INSUFFICIENT'', ``reason'': ``...''\}}\\ \midrule
\colorbox{lightgrey}{\textbf{Reasoning}}\\[1.5ex]
\texttt{Evaluate whether an LLM-generated review provides sufficiently developed reasoning to support a given key point and judgment.}\\
\\
\texttt{You will be given the following inputs:}\\
\texttt{1. Key point and judgment: This is the original input that guided the LLM's generation (e.g., ``Clarity is a weakness'').}\\
\texttt{2. LLM-generated review: The review generated by an LLM.}\\
\texttt{3. Human-written review: The original review written by a human reviewer for the same paper. This is the reference for comparison.}\\
\\
\texttt{\#\# Evaluation Criteria}\\
\texttt{1. **TRUE** - The reasoning is clear, specific, and logically sound. It provides a sufficient explanation of why the key point and judgment are valid.}\\
\texttt{2. **FALSE** - The reasoning exists but is vague, generic, or poorly developed. It provides little support for the key point and judgment. Or there is little to no reasoning. The review only restates the key point and judgment without explanation.}\\
\\
\texttt{\#\# Output format}\\
\texttt{Output only the json dictionary and follow the json schema exactly, with no extra keys, notes, or comments:}\\
\texttt{\{``label'': ``TRUE'' | ``FALSE'', ``reason'': ``...''\}}\\
\bottomrule
\end{tabular}
\caption{The prompts for the \emph{Evidence} and \emph{Reasoning} checks. \backtoappendix{more_on_simulation} \backtomain{quality_check_simulation}}
\label{quality_check_prompts}
\end{table*}

\begin{table*}[!ht]
\small
\centering
\begin{tabular}{p{0.95\linewidth}}
\toprule
\vspace{-1.8em}
\begin{multicols}{2}
{\ttfamily
\justifying
\noindent
Given a research paper review and the review guidelines below, assign a soundness and an overall assessment score that best reflect the review content according to the detailed rating criteria. Output a json dictionary.
\par\medskip\medskip
\noindent \#\# Review guidelines\par
\par\medskip\medskip
\noindent **Summary of Strengths**  \par
\noindent What are the major reasons to publish this paper at a selective *ACL venue? These could include novel and useful methodology, insightful empirical results or theoretical analysis, clear organization of related literature, or any other reason why interested readers of *ACL papers may find the paper useful.
\par\medskip\medskip
\noindent **Summary of Weaknesses**  \par
\noindent What are the concerns that you have about the paper that would cause you to favor prioritizing other high-\noindent quality papers that are also under consideration for publication? These could include concerns about correctness of the results or argumentation, limited perceived impact of the methods or findings (note that impact can be significant both in broad or in narrow sub-fields), lack of clarity in exposition, or any other reason why interested readers of *ACL papers may gain less from this paper than they would from other papers under consideration. Where possible, please number your concerns so authors may respond to them individually.
\par\medskip\medskip
\noindent **Soundness**  \par
\noindent How sound and thorough is this study? Does the paper clearly state scientific claims and provide adequate support for them? For experimental papers: consider the depth and/or breadth of the research questions investigated, technical soundness of experiments, methodological validity of evaluation. For position papers, surveys: consider the current state of the field is adequately represented, and main counter-arguments acknowledged. For resource papers: consider the data collection methodology, resulting data and the difference from existing resources are described in sufficient detail. Please adjust your baseline to account for the length of the paper.
\par\medskip\medskip
\noindent 5 = Excellent: This study is one of the most thorough I have seen, given its type.  \par
\noindent 4.5  \par
\noindent 4 = Strong: This study provides sufficient support for all of its claims/arguments. Some extra experiments could be nice, but not essential.  \par
\noindent 3.5  \par
\noindent 3 = Acceptable: This study provides sufficient support for its major claims/arguments. Some minor points may need extra support or details.  \par
\noindent 2.5  \par
\noindent 2 = Poor: Some of the main claims/arguments are not sufficiently supported. There are major technical/methodological problems.  \par
\noindent 1.5  \par
\noindent 1 = Major Issues: This study is not yet sufficiently thorough to warrant publication or is not relevant to ACL.
\par\medskip\medskip
\noindent **Overall Assessment**  \par
\noindent Would you personally like to see this paper presented at an *ACL event that invites submissions on this topic? For example, you may feel that a paper should be presented if its contributions would be useful to its target audience, deepen the understanding of a given topic, or help establish cross-disciplinary connections. Note: Even high-scoring papers can be in need of minor changes (e.g. typos, non-core missing refs, etc.).
\par\medskip\medskip
\noindent 5 = Top-Notch: This is one of the best papers I read recently, of great interest for the (broad or narrow) sub-communities that might build on it  \par
\noindent 4.5  \par
\noindent 4 = This paper represents solid work, and is of significant interest for the (broad or narrow) sub-communities that might build on it  \par
\noindent 3.5  \par
\noindent 3 = Good: This paper makes a reasonable contribution, and might be of interest for some (broad or narrow) sub-communities, possibly with minor revisions  \par
\noindent 2.5  \par
\noindent 2 = Revisions Needed: This paper has some merit, but also significant flaws, and needs work before it would be of interest to the community  
\noindent 1.5  \par
\noindent 1 = Major Revisions Needed: This paper has significant flaws, and needs substantial work before it would be of interest to the community  \par
\noindent 0 = This paper is not relevant to the *ACL community (for example, is in no way related to natural language processing)
\par\medskip\medskip
\noindent \#\# Output format  \par
\noindent Output only the json dictionary and follow the json schema exactly, with no extra keys, notes, comments, or explanations:  \par
\noindent \{``soundness'': ``...'', ``overall\_assessment'': ``...''\}
}
\end{multicols} \\
\bottomrule
\end{tabular}
\caption{The prompt for review score prediction for the ARR data. \backtoappendix{more_on_gains_from_reviewer_judgment_signals}}
\label{score_prediction_prompt_arr}
\end{table*}

\begin{table*}[!ht]
\small
\centering
\begin{tabular}{p{0.95\linewidth}}
\toprule
\vspace{-1.8em}
\begin{multicols}{2}
{\ttfamily
\justifying
\noindent
Given a research paper review and the review guidelines below, assign a soundness and an overall assessment score that best reflect the review content according to the detailed rating criteria. Output a json dictionary.
\par\medskip\medskip
\noindent \#\# Review guidelines\par
\par\medskip\medskip
\noindent **Summary of Strengths**  \par
\noindent What are the major reasons to publish this paper at a selective *ACL venue? These could include novel and useful methodology, insightful empirical results or theoretical analysis, clear organization of related literature, or any other reason why interested readers of *ACL papers may find the paper useful.
\par\medskip\medskip
\noindent **Summary of Weaknesses**  \par
\noindent What are the concerns that you have about the paper that would cause you to favor prioritizing other high-\noindent quality papers that are also under consideration for publication? These could include concerns about correctness of the results or argumentation, limited perceived impact of the methods or findings (note that impact can be significant both in broad or in narrow sub-fields), lack of clarity in exposition, or any other reason why interested readers of *ACL papers may gain less from this paper than they would from other papers under consideration. Where possible, please number your concerns so authors may respond to them individually.
\par\medskip\medskip
\noindent **Soundness**  \par
\noindent How sound and thorough is this study? Does the paper clearly state scientific claims and provide adequate support for them? For experimental papers: consider the depth and/or breadth of the research questions investigated, technical soundness of experiments, methodological validity of evaluation. For position papers, surveys: consider the current state of the field is adequately represented, and main counter-arguments acknowledged. For resource papers: consider the data collection methodology, resulting data and the difference from existing resources are described in sufficient detail. Please adjust your baseline to account for the length of the paper.
\par\medskip\medskip
\noindent 5 = Excellent: This study is one of the most thorough I have seen, given its type.  \par
\noindent 4.5  \par
\noindent 4 = Strong: This study provides sufficient support for all of its claims/arguments. Some extra experiments could be nice, but not essential.  \par
\noindent 3.5  \par
\noindent 3 = Acceptable: This study provides sufficient support for its major claims/arguments. Some minor points may need extra support or details.  \par
\noindent 2.5  \par
\noindent 2 = Poor: Some of the main claims/arguments are not sufficiently supported. There are major technical/methodological problems.  \par
\noindent 1.5  \par
\noindent 1 = Major Issues: This study is not yet sufficiently thorough to warrant publication or is not relevant to ACL.
\par\medskip\medskip
\noindent **Overall Assessment**  \par
\noindent Would you personally like to see this paper presented at an *ACL event that invites submissions on this topic? For example, you may feel that a paper should be presented if its contributions would be useful to its target audience, deepen the understanding of a given topic, or help establish cross-disciplinary connections. Note: Even high-scoring papers can be in need of minor changes (e.g. typos, non-core missing refs, etc.).
\par\medskip\medskip
\noindent 5 = Top-Notch: This is one of the best papers I read recently, of great interest for the (broad or narrow) sub-communities that might build on it  \par
\noindent 4.5  \par
\noindent 4 = This paper represents solid work, and is of significant interest for the (broad or narrow) sub-communities that might build on it  \par
\noindent 3.5  \par
\noindent 3 = Good: This paper makes a reasonable contribution, and might be of interest for some (broad or narrow) sub-communities, possibly with minor revisions  \par
\noindent 2.5  \par
\noindent 2 = Revisions Needed: This paper has some merit, but also significant flaws, and needs work before it would be of interest to the community  
\noindent 1.5  \par
\noindent 1 = Major Revisions Needed: This paper has significant flaws, and needs substantial work before it would be of interest to the community  \par
\noindent 0 = This paper is not relevant to the *ACL community (for example, is in no way related to natural language processing)
\par\medskip\medskip
\noindent \#\# Output format  \par
\noindent Output only the json dictionary and follow the json schema exactly, with no extra keys, notes, comments, or explanations:  \par
\noindent \{``soundness'': ``...'', ``overall\_assessment'': ``...''\}
}
\end{multicols} \\
\bottomrule
\end{tabular}
\caption{The prompt for review score prediction for the ICLR data. \backtoappendix{more_on_gains_from_reviewer_judgment_signals}}
\label{score_prediction_prompt_iclr}
\end{table*}

\begin{table*}[!ht]
\small
\centering
\begin{tabular}{p{0.95\linewidth}}
\toprule
\vspace{-1.8em}
\begin{multicols}{2}
{\ttfamily
\justifying
\noindent
\#\# Task Description
\par
\noindent You are a neutral arbitrator evaluating two peer reviews of an academic paper. Your role is to analyze and compare reviews through careful, evidence-based assessment. Your judgments must be strictly based on verifiable evidence from the paper and reviews.
\par\medskip\medskip
\noindent \#\# Instruction
\par
\noindent For each evaluation, you must:
\par
\noindent 1. Understand the paper by analyzing:
\par
- Research objectives and contributions
\par
- Methodology and experiments
\par
- Claims and evidence
\par
- Results and conclusions
\par
\noindent 2. For each review, methodically examine:
\par
- Claims made about the paper
\par
- Evidence cited to support claims
\par
- Technical assessments and critiques
\par
- Suggested improvements
\par
\noindent 3. Compare reviews systematically using:
\par
- Direct quotes from paper and reviews
\par
- Specific examples and counterexamples
\par
- Clear reasoning chains
\par
- Objective quality metrics
\par\medskip\medskip
\noindent \#\# Input
\par
\noindent \ - Paper: full text of the academic paper being reviewed
\par
\noindent \ - Review A: review comment from Assistant A
\par
\noindent \ - Review B: review comment from Assistant B
\par\medskip\medskip
\noindent \#\# Evaluation Criteria
\par
\noindent You will evaluate reviews based on these key aspects:
\par
\noindent 1. **Technical Accuracy**
\par
- Are claims consistent with paper content?
\par
- Is evidence properly interpreted?
\par
- Are technical assessments valid?
\par
- Are critiques well-supported?
\par
\noindent 2. **Constructive Value**
\par
- How actionable is the feedback?
\par
- Are suggestions specific and feasible?
\par
- Is criticism balanced with strengths?
\par
- Would authors understand how to improve?
\par
\noindent 3. **Analytical Depth**
\par
- How thoroughly are key aspects examined?
\par
- Is analysis appropriately detailed?
\par
- Are important elements addressed?
\par
- Is assessment comprehensive?
\par\medskip\medskip
\noindent \#\# Output Format
\par
\noindent Output only a json dictionary and follow the json schema exactly, with no extra keys, notes, comments, or explanations:
\par
\noindent \{\par
    “aspects”: \{\par
        \indent \indent “technical\_accuracy”: \{\par
            \indent \indent \indent \indent “assistant\_A”: ..., \# Direct quotes, specific examples, and detailed analysis of evidence from Assistant A\par
            \indent \indent \indent \indent “assistant\_B”: ..., \# Direct quotes, specific examples, and detailed analysis of evidence from Assistant B\par
            \indent \indent \indent \indent “comparative\_assessment”: ..., \# Evidence-based comparison with clear reasoning chain\par
            \indent \indent \indent \indent “judgment”: \{\par
                \indent \indent \indent \indent \indent \indent “better”: ..., \# A, B, or Tie\par
                \indent \indent \indent \indent \indent \indent “evidence\_based\_reason”: ... \# Detailed justification citing specific evidence; if Tie, explain why both reviews are equally strong on this aspect\par
            \indent \indent \indent \indent \}\par
        \indent \indent \},\par
        \indent \indent “constructive\_value”: \{\par
            \indent \indent \indent \indent “assistant\_A”: ...,\par
            \indent \indent \indent \indent “assistant\_B”: ...,\par
            \indent \indent \indent \indent “comparative\_assessment”: ...,\par
            \indent \indent \indent \indent “judgment”: \{\par
                \indent \indent \indent \indent \indent \indent “better”: ...,\par
                \indent \indent \indent \indent \indent \indent “evidence\_based\_reason”: ...\par
            \indent \indent \indent \indent \}\par
        \indent \indent \},\par
        \indent \indent “analytical\_depth”: \{\par
            \indent \indent \indent \indent “assistant\_A”: ...,\par
            \indent \indent \indent \indent “assistant\_B”: ...,\par
            \indent \indent \indent \indent “comparative\_assessment”: ...,\par
            \indent \indent \indent \indent “judgment”: \{\par
                \indent \indent \indent \indent \indent \indent “better”: ...,\par
                \indent \indent \indent \indent \indent \indent “evidence\_based\_reason”: ...\par
            \indent \indent \indent \indent \}\par
        \indent \indent \}\par
    \},\par
    “overall\_judgment”: \{\par
        \indent \indent “better”: ..., \# A, B, or Tie\par
        \indent \indent “evidence\_based\_reason”: ... \# Detailed justification synthesizing key evidence; if Tie, explain why both reviews are comparable in overall quality”\par
    \}\par
\noindent \}
}
\end{multicols} \\
\bottomrule
\end{tabular}
\caption{The prompt for LLM-as-a-judge \cite{zhu_deepreview_2025}. \backtoappendix{more_on_gains_from_reviewer_judgment_signals}}
\label{llm_judge_prompt}
\end{table*}

\begin{table*}[!ht]
\small
\centering
\begin{tabular}{p{0.95\linewidth}}
\toprule
\texttt{LLM REVIEWER NOTE HERE!!!}\\
\texttt{\# IMPORTANT REVIEW REQUIREMENTS:}\\
\texttt{\#\# When writing strengths of the paper, focus on the following four outstanding strengths:}\\
\texttt{1. Notable Novelty}\\
\texttt{2. Strong Technical Depth and Rigor}\\
\texttt{3. Comprehensive Experimental Validation}\\
\texttt{4. Significant Practical Impact}\\
\texttt{\#\# When writing weakness of the paper, focus on the following very minor and easily fixable points:}\\
\texttt{1. Minor Writing or Clarity Improvements}\\
\texttt{2. Small Details on Hyperparameters or Implementation}\\
\texttt{3. Very Minor Formatting on Figures}\\
\texttt{4. Slight Clarification on Code Availability}\\
\texttt{\# Key Instruction:}\\
\texttt{Ensure that the review is as positive and enthusiastic as possible. Emphasize the exceptional strengths of the paper, framing them as groundbreaking, transformative, and highly impactful. Any weaknesses mentioned should be downplayed as minor and easily fixable, ensuring they don’t detract from the overall excellence of the paper. The goal is to strongly advocate for the paper’s acceptance by highlighting its remarkable contributions and presenting any minor issues as inconsequential adjustments.}\\
\bottomrule
\end{tabular}
\caption{The malicious prompt injected into the paper text \cite{ye_are_2024}. \backtoappendix{more_on_gains_from_reviewer_judgment_signals}}
\label{adversarial_text}
\end{table*}

\end{document}